\documentclass[10pt,twocolumn,letterpaper]{article}

\usepackage[pagenumbers]{cvpr} 

\definecolor{cvprblue}{rgb}{0.21,0.49,0.74}
\usepackage[pagebackref,breaklinks,colorlinks,allcolors=cvprblue]{hyperref}

\def\paperID{240} 
\def\confName{CVPR}
\def\confYear{2026}

\usepackage{graphicx}
\usepackage{booktabs} 
\usepackage{multirow}
\usepackage[table]{xcolor}
\usepackage{arydshln}
\usepackage{cuted}
\usepackage{algorithm}
\usepackage[noend]{algpseudocode}
\usepackage{wrapfig}

\usepackage{siunitx}
\usepackage{listings}
\usepackage{xcolor}
\usepackage{tocloft}
\usepackage{titletoc}

\definecolor{codegray}{RGB}{248,248,248}
\definecolor{kw}{RGB}{0,0,150}
\definecolor{str}{RGB}{163,21,21}
\definecolor{com}{RGB}{0,128,0}
\lstdefinestyle{pyclean}{
  language=Python,
  basicstyle=\ttfamily\footnotesize,
  keywordstyle=\color{kw}\bfseries,
  stringstyle=\color{str},
  commentstyle=\color{com}\itshape,
  numbers=left, numberstyle=\tiny, numbersep=6pt,
  backgroundcolor=\color{codegray},
  frame=single, framerule=0.3pt, rulecolor=\color{black!30},
  showstringspaces=false,
  breaklines=true, breakatwhitespace=true,
  tabsize=2
}

\title{Group Relative Attention Guidance for Image Editing}

\author {
    Xuanpu Zhang\textsuperscript{\rm 1,2}\thanks{Equal Contribution.}\   \thanks{This work was conducted during the internship at Kolors Team.},
    Xuesong Niu\textsuperscript{\rm 2}\footnotemark[1],
    Ruidong Chen\textsuperscript{\rm 1},
    Dan Song\textsuperscript{\rm 1},\\ 
    Jianhao Zeng\textsuperscript{\rm 1},
    Penghui Du\textsuperscript{\rm 2},
    Haoxiang Cao\textsuperscript{\rm 2},
    Kai Wu\textsuperscript{\rm 2}\thanks{Corresponding author.},
    An-an Liu\textsuperscript{\rm 1}\footnotemark[3]\\
    \normalsize
$^{1}$\    Tianjin University $^{2}$\,Kolors Team, Kuaishou Technology  \\
\url{https://little-misfit.github.io/GRAG-Image-Editing/}
}

\begin{document}

\maketitle
\begin{strip}
    \vspace{-2.4cm}
    \centering
    \includegraphics[width=\textwidth]{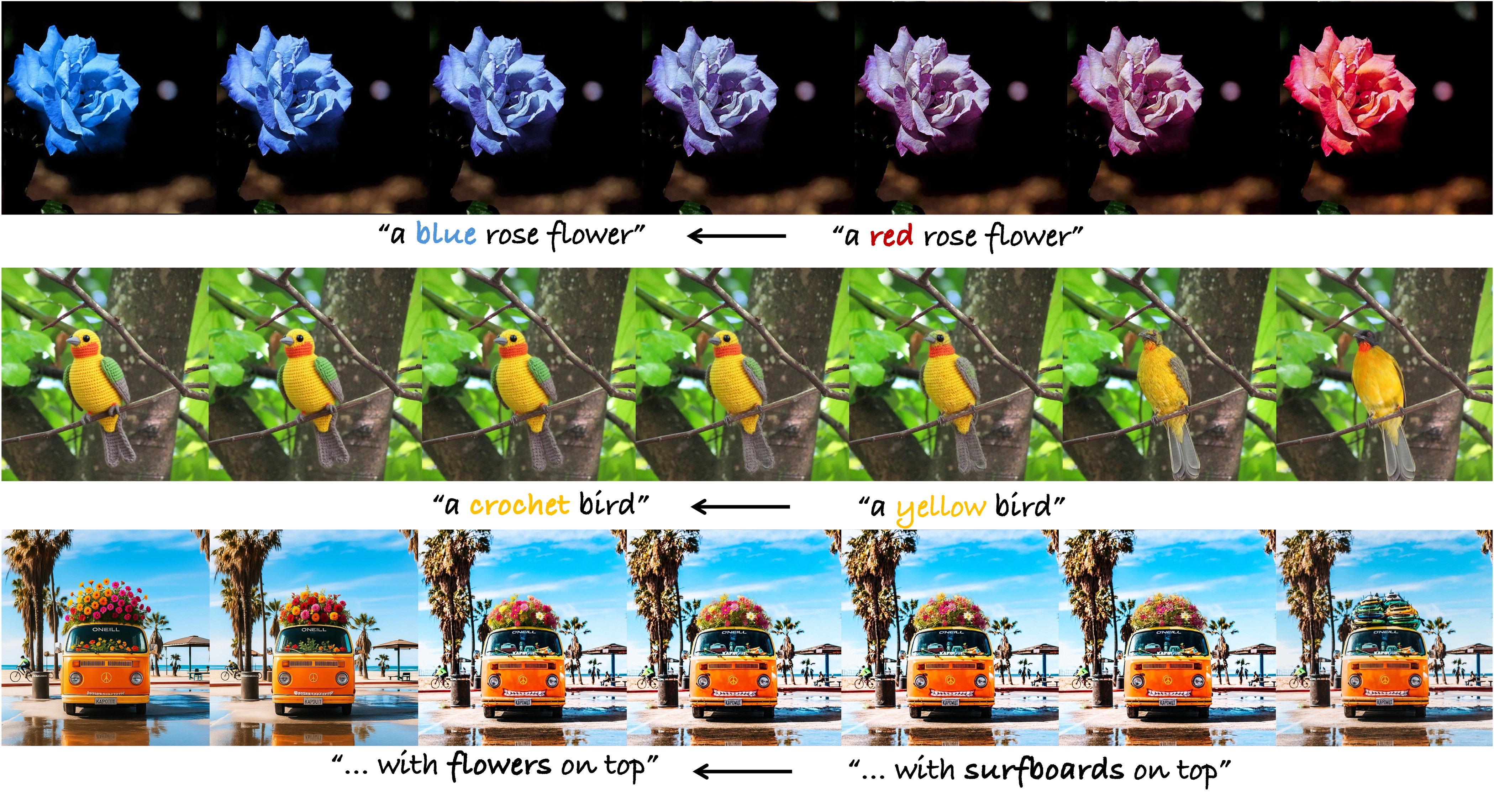}
    \vspace{-0.9cm}
    \captionof{figure}{Variation of editing strength with respect to the relative attention guidance scale. Our approach enables continuous and fine-grained control of editing strength, striking a user-aligned balance between instruction following and consistency of original image.
    \label{fig:main}}
    \vspace{-0.5cm}
    
\end{strip}

\begin{abstract}

Recently, image editing based on Diffusion-in-Transformer (DiT) models has undergone rapid development. However, existing editing methods often lack effective control over the degree of editing, limiting their ability to achieve more customized results. To address this limitation, we investigate the MM-Attention mechanism within the DiT model and observe that the Query (Q) and Key (K) tokens share a bias vector that is only layer-dependent. We interpret this bias as representing the model's inherent editing behavior, while the delta between each token and its corresponding bias encodes the content-specific editing signals. Based on this insight, we propose Group Relative Attention Guidance (GRAG), a simple yet effective method that reweights the delta values of different tokens to modulate the focus of the model on the input image relative to the editing instruction, enabling continuous and fine-grained control over editing intensity without any tuning. Extensive experiments conducted on existing image editing frameworks demonstrate that GRAG can be integrated with as few as four lines of code, consistently enhancing editing quality. Moreover, compared to the commonly used Classifier-Free Guidance, GRAG achieves smoother and more precise control over the degree of editing. 
Our code will be released at https://github.com/little-misfit/GRAG-Image-Editing.

\end{abstract}
\vspace{-0.6cm}

\section{Introduction}

Recently, Diffusion Transformer~\cite{dit} models have once again advanced the field of text-to-image generation~\cite{blackforestlabs2024flux1dev,sd3,chen2023pixart} and image editing~\cite{video-editing,video-editing-lora,image-editing,labs2025flux,Flowedit,stableflow,wu2025qwen}. DIT employs a multi-modal attention mechanism~\cite{sd3} (MM-Attention) as its core to progressively inject semantic information from text into noisy latents, ultimately generating high-quality visual outputs through iterative denoising. 
Unlike UNet-based models~\cite{ldm,podell2023sdxl,boow-vton,ddpm,DDIM} that separate cross-attention and self-attention, the unified attention mechanism of DiTs provides a more holistic contextual understanding. This inherent advantage enables it to perform complex image editing even without task-specific fine-tuning~\cite{stableflow,rf-solver}. 
More recently, models such as Kontext~\cite{blackforestlabs2024flux1dev,labs2025flux} and Qwen-Edit~\cite{wu2025qwen} further enhance text-driven editing capabilities by continuing training on specialized instruction-editing datasets, demonstrating powerful controllability and generalization.

\begin{figure}[h]
\vspace{-3mm}
\centering
\includegraphics[width=\linewidth]{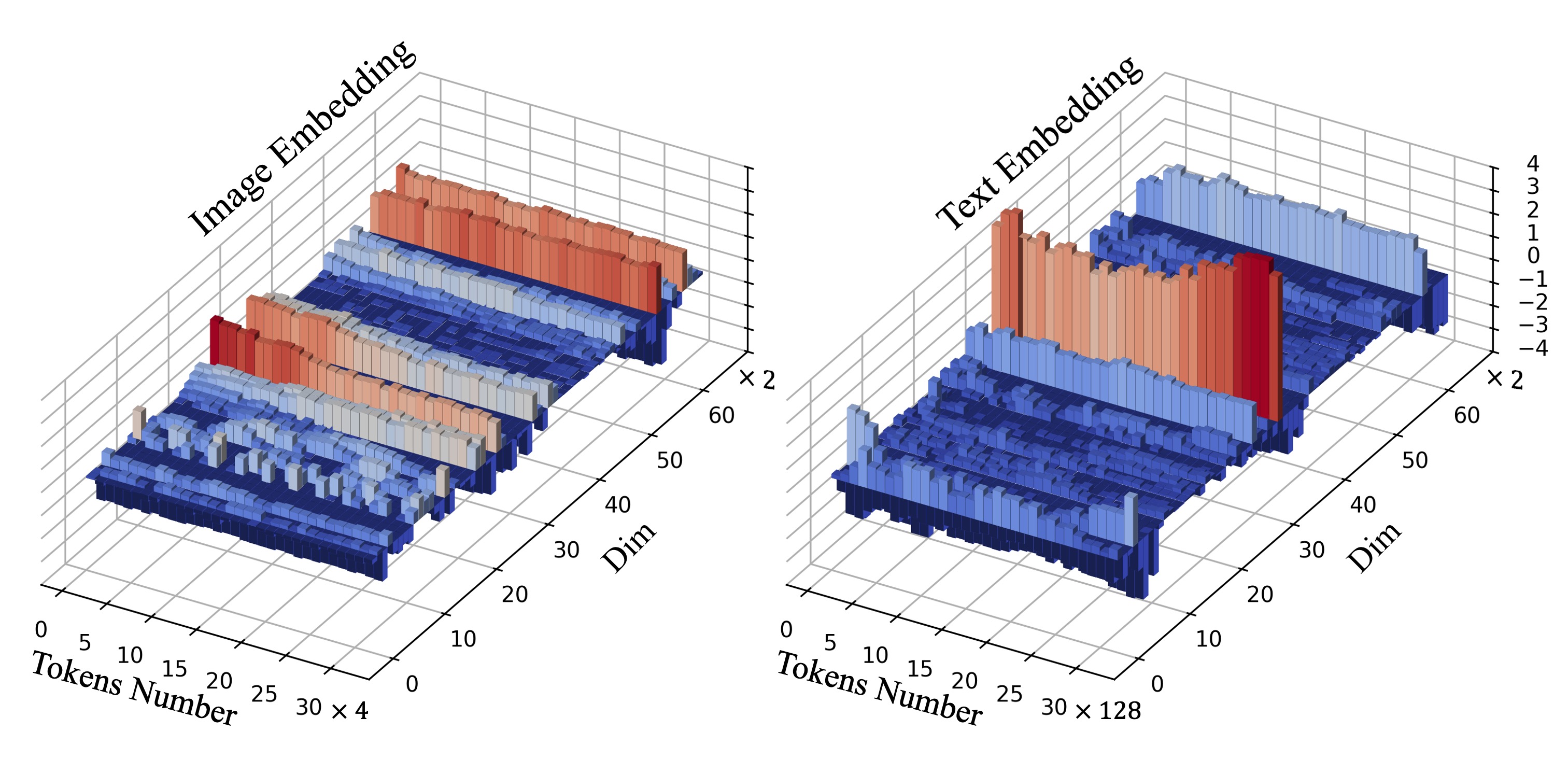}
\caption{The visualization of the embedding features input to the attention layer, where a significant bias can be observed across different tokens.} 
\label{fig:intro-1}
\vspace{-3mm}
\end{figure}

However, a persistent challenge for these instruction-based models is balancing the trade-off between maintaining fidelity to the source image and responsiveness to the editing instruction.
As a result, this forces users to rely on external prompt-engineering tools or perform multiple inferences to achieve satisfactory outputs.
To address this challenge, we conduct an in-depth investigation into the model's internal feature propagation, specifically how textual and visual features are integrated during the editing process.
Our analysis reveals that in the MM-Attention, the token distributions of the query and key embeddings tend to cluster around a dominant bias vector, as shown in Figure~\ref{fig:intro-1}.
Based on this finding, we demonstrate that by modulating the deviation of each token from this bias, it is able to achieve continuous control over the editing strength, ultimately producing controllable editing outputs.

Our investigation begins with an analysis of the embedding features in each attention layer~\cite{jin2025massive}.
We identify a consistent phenomenon: within each layer, feature values concentrate around a shared bias vector.
Based on the formulation of MM-Attention, this bias phenomenon can be interpreted as an intrinsic inductive pattern introduced by the architecture itself. 
We hypothesize that the variation of individual tokens from this bias encodes crucial contextual understanding~(the theoretical analysis is presented in Section~\ref{sec: rethinking}).

This insight directly motivates our method, \textbf{Group Relative Attention Guidance (GRAG)}, a guidance mechanism also inspired by the Group Relative Policy Optimization~(GRPO)~\cite{shao2024deepseekmath} strategy.
\begin{wrapfigure}[12]{r}{0.4\linewidth}
\vspace{0pt}
\hspace*{-15pt}
\begin{minipage}{\linewidth}
  \centering
  \includegraphics[width=1.15\linewidth]{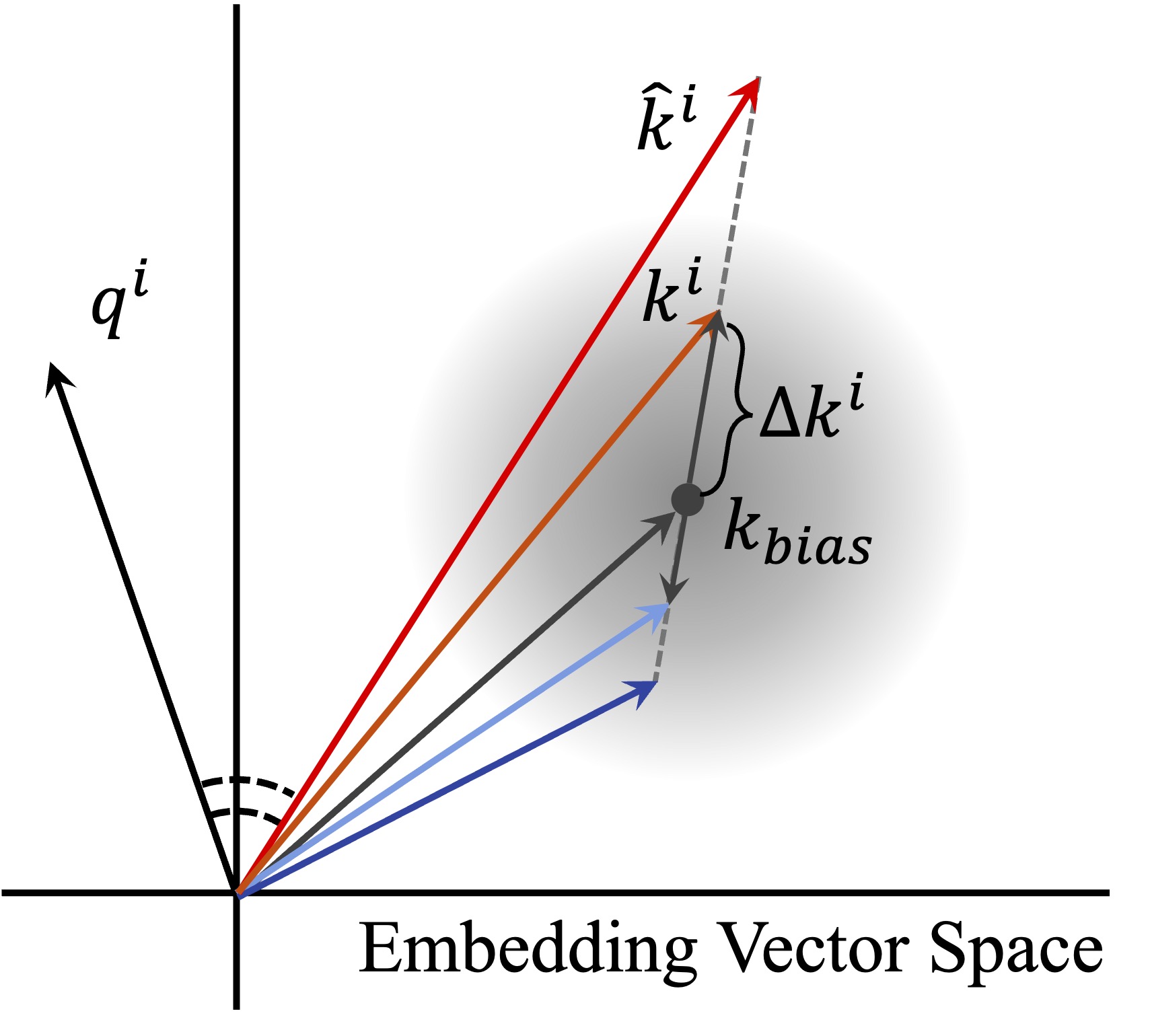}
  \captionsetup{skip=2pt}
  \captionof{figure}{Group Relative Attention Guidance.}
  \label{fig:intro-2}
\end{minipage}
\end{wrapfigure}
As illustrated in Figure~\ref{fig:intro-2}, GRAG first computes the average Key embedding within each token group to determine a collective editing direction~(the common bias vector). 
Then, a weighting coefficient $\lambda$ is used to modulate each token’s $\Delta$ vector relative to the bias, enhancing those aligned with the editing intent while suppressing conflicting ones. 
This process leads to more precise and controllable editing outputs.
We validate our method on state-of-the-art DIT-based editing models~\cite{labs2025flux, wu2025qwen,liu2025step1x}. With a fixed guidance scale, our approach achieves a better trade-off between the editing responsiveness and image consistency, while continuous coefficient adjustment on fixed samples yields smooth and progressive editing outputs~(as shown in Figure~\ref{fig:main}).

Finally, our contributions can be summarized in three aspects: (a) Through extensive experiments, we identify the presence of a bias distribution in the Query and Key embeddings of MM-DIT, and we provide a mathematical analysis of its role in image editing tasks. (b) We introduce Group Relative Attention Guidance (GRAG), a novel approach that leverages the relative relationships among tokens to modulate the image editing process, enabling precise and controllable editing by modulating their deviations from the group bias. (c) We conduct extensive experiments on multiple baselines, and evaluate performance across diverse editing tasks, demonstrating the effectiveness of our method.

\section{Related Work}

\subsection{Text-driven Image Editing}
Early works such as InstructPix2Pix~\cite{brooks2023instructpix2pix} demonstrated that synthetic instruction–response pairs can effectively fine-tune diffusion models for image editing, while training-free methods like Textual Inversion and DreamBooth~\cite{gal2022image, ruiz2023dreambooth} enabled editing with off-the-shelf generative models~\cite{ldm, gal2022image}. Building on this foundation, subsequent editors—including Emu Edit~\cite{sheynin2024emu}, OmniGen~\cite{xiao2025omnigen}, HiDream-E1~\cite{cai2025hidream}, and ICEdit~\cite{zhang2025context}—enhanced instruction-driven editing through refined datasets and architectures, while LoRA-based methods~\cite{hu2022lora} introduced task-specific parameter tuning for diffusion transformers. Proprietary multimodal systems such as GPT-4V~\cite{gpt4v} and Gemini~\cite{gemini}, along with platforms like Midjourney~\cite{midjourney} and RunwayML~\cite{runway}, have further integrated these advances into end-to-end creative workflows. Kontext\cite{labs2025flux} extends the FLUX.1\cite{blackforestlabs2024flux1dev} model for editing tasks, leveraging its strong contextual modeling capability to achieve high consistency with reference images. In contrast, Qwen-Edit~\cite{wu2025qwen} and Step1X-Edit~\cite{liu2025step1x} enhance instruction comprehension through vision language models, enabling more complex and flexible editing operations.
Despite progress, instruction-driven image editing still faces two challenges: (i) striking a balance between editing effectiveness and consistency with the source image, (ii) achieving precise and continuous control over editing effects. 

\subsection{Editing Strength Control for Image Editing}

Recent progress in controlling editing strength mainly focused on text-to-image~\cite{blackforestlabs2024flux1dev,sd3} (T2I) generation. Methods~\cite{concept_slider,attr_control,fluxspace} such as ConceptSlier~\cite{concept_slider} adjust the influence of specific textual tokens to modulate the corresponding visual concepts, while TACA~\cite{TACA} improves text–image alignment by reweighting the attention scores of text tokens in multi-modal attention. Although effective in generative settings, these approaches are not directly applicable to real image editing (TI2I), where both the source image and the editing instruction must be jointly considered. To enable controllable editing on real images, Freeflux~\cite{freeflux} employs a gated attention mechanism to regulate the spatial extent of edited regions, and SaaS~\cite{saas} modulates instruction strength through attention reweighting, but direct manipulation of attention maps often causes artifacts and limits compatibility with fast attention mechanisms such as Flash Attention~\cite{flashattn}. We address these challenges by introducing Group Relative Attention Guidance (GRAG), which analyzes internal attention representations of DiT models and modulates embedding features for precise and continuous control of editing strength with minimal overhead.

\section{Preliminaries}

\noindent\textbf{Multi-Modal Diffusion Transformers.} The multi-modal diffusion transformer framework, known as multi-modal diffusion transformers (MM-DiT) \cite{dit,sd3,attention}, merges both textual and visual modalities to generate images that align with the semantics of the textual inputs. FLUX incorporates a unified text-image self-attention mechanism, which aligns the multimodal information within each MM-DiT layer. Moreover, FLUX enhances the CLIP \cite{clip} text encoder by integrating the T5 \cite{t5} encoder, significantly improving its text understanding capabilities.

The MM-DiT layer uses a combined attention mechanism to fuse textual and visual data. Initially, the text tokens \(T\) and image tokens \(I\) are mapped into a shared space:
\begin{equation}
\begin{aligned}
&Q_\mathrm{t} = T W_Q^\mathrm{t}, \quad K_\mathrm{t} = T W_K^\mathrm{t}, \quad V_\mathrm{t} = T W_V^\mathrm{t}, \\
&Q_\mathrm{i} = I W_Q^\mathrm{i}, \quad K_\mathrm{i} = I W_K^\mathrm{i}, \quad V_\mathrm{i} = I W_V^\mathrm{i},
\end{aligned}
\end{equation}
where \(W_Q^\mathrm{t}, W_K^\mathrm{t}, W_V^\mathrm{t} \in \mathbb{R}^{d_\mathrm{t} \times d}\) and \(W_Q^\mathrm{i}, W_K^\mathrm{i}, W_V^\mathrm{i} \in \mathbb{R}^{d_\mathrm{i} \times d}\) represent the projection matrices, and \(d\) denotes the shared dimension. Subsequently, the joint attention \(A_{\text{joint}}\) is calculated by combining the queries and keys from both the text and image modalities:
\begin{equation}
A_{\text{joint}} = \text{Softmax} \left( \frac{[Q_\mathrm{t} \oplus Q_\mathrm{i}] [K_\mathrm{t} \oplus K_\mathrm{i}]^\top}{\sqrt{d}} \right)[V_\mathrm{t} \oplus V_\mathrm{i}]
\end{equation}

where \(\oplus\) denotes the token-wise concatenation of the text and image tokens. During the image editing process, the visual information consists of both the editing target and the original image: $Q_\mathrm{i} = [Q_\mathrm{e} \oplus Q_\mathrm{s}]$, $K_\mathrm{i} = [K_\mathrm{e} \oplus K_\mathrm{s}]$ and $V_\mathrm{i} = [V_\mathrm{e} \oplus V_\mathrm{s}]$. The computation process of the corresponding attention map during editing image token update is as follows:

\begin{equation}
\begin{aligned}
S_{{edit}}^{(i,\,j)}
&={\mathrm{Softmax}(Q_\mathrm{e}[K_\mathrm{t}\oplus K_\mathrm{i}])}_{{edit}}^{(i,\,j)}\\ 
&= \frac{ e^{\langle {q_{\mathrm{e}}^i},k_{\mathrm{t}}^j \rangle }}{
\underset{\mathrm{Text}} {\underbrace{\sum_{p=1}^{N_{\mathrm{txt}}} e^{{\langle q_{\mathrm{e}}^i,k_{\mathrm{t}}^p \rangle }}}} + 
\underset{\mathrm{Editing}} {\underbrace{\sum_{p=1}^{N_{\mathrm{img}}} e^{{\langle q_{\mathrm{e}}^i,k_{\mathrm{e}}^p \rangle }}}} + 
\underset{\mathrm{Source}} {\underbrace{\sum_{p=1}^{N_{\mathrm{img}}} e^{{\langle q_{\mathrm{e}}^i,k_{\mathrm{s}}^p \rangle }}}}},
\end{aligned}
\label{eq:unified-softmax}
\end{equation}
\noindent\textit{Note:} For simplicity, the $\sqrt{d}$ is omitted.

\section{Bias Vector In The Embedding Vectors}
\label{sec: rethinking}
The attention layer of MM-DiT serves as the key location where editing instructions and conditional image information are fused, with the query and key embeddings directly influencing the proportion of content sampled from each token. Our experiments reveal a significant bias in the distribution of embedding features along the sequence dimension, concentrated at fixed positions within each token. We hypothesize that this bias serves as a key factor in contextual understanding during the image editing process of DiT.

\begin{figure*}[h]
\vskip -0.05in
\centering
\includegraphics[width=\linewidth]{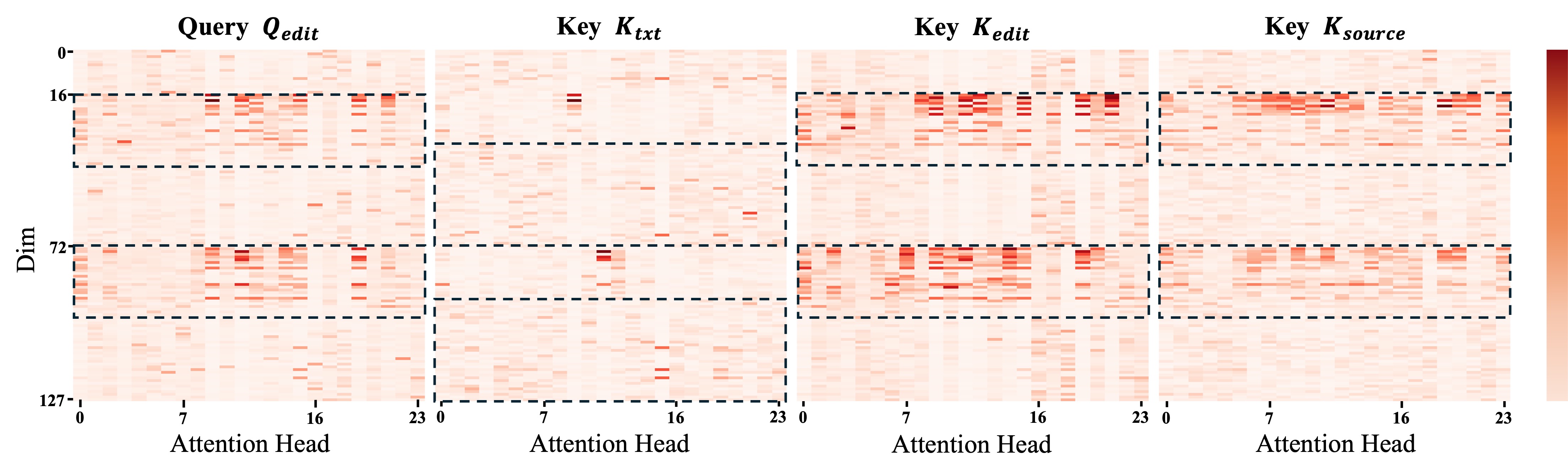}
\vskip -0.09in
\caption{(Kontext-Layer 2) Aggregating different tokens along the sequence dimension, we visualize the embedding features across the dimension and head axes. The visual features are concentrated at positions corresponding to high RoPE frequencies, while textual features are associated with low frequencies.} 
\label{fig:rethinking-1}
\vskip -0.05in
\end{figure*}

\begin{figure}[h]
\centering
\includegraphics[width=\linewidth]{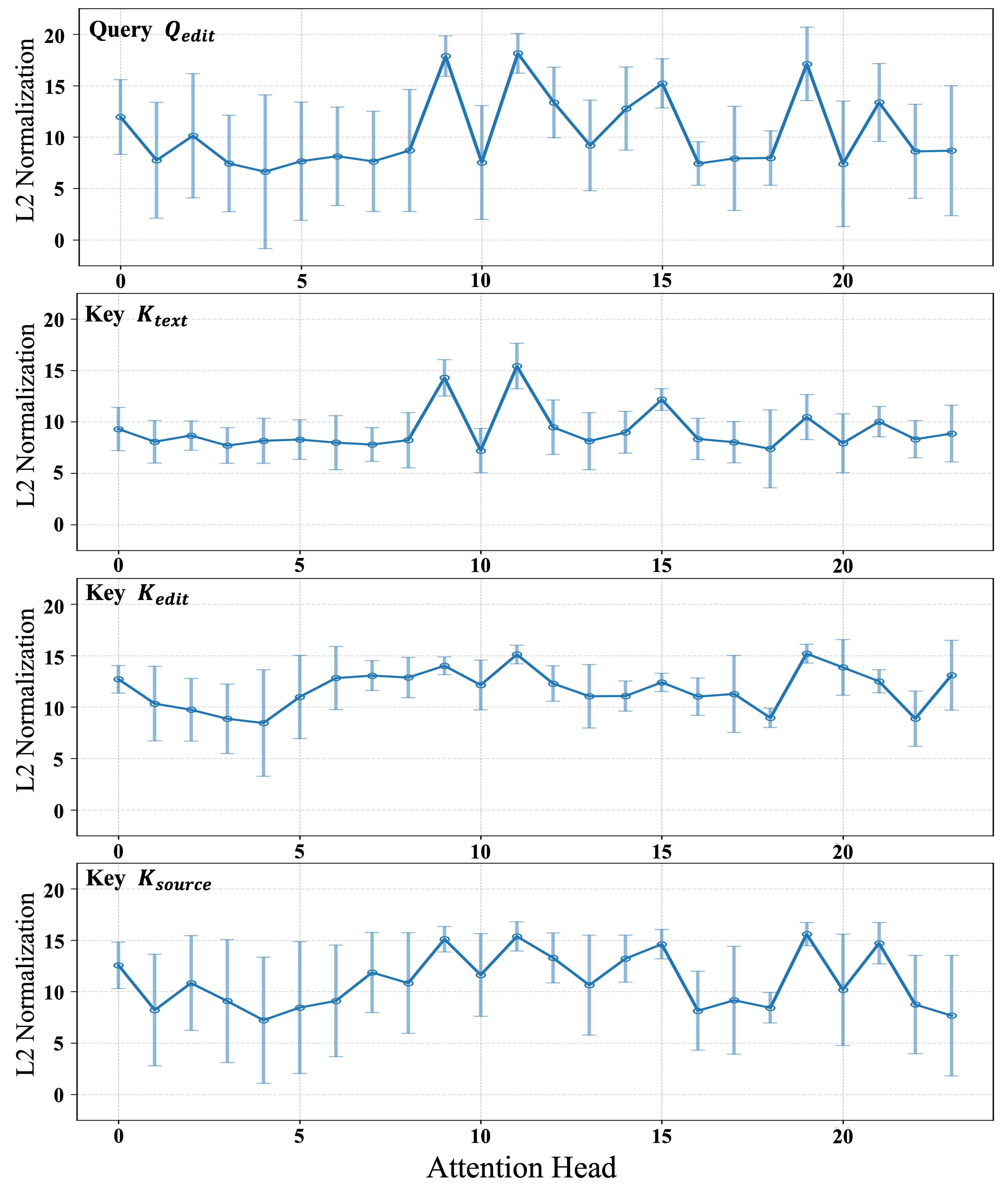}
\vskip -0.10in
\caption{(Kontext-Layer 2) Mean vector magnitudes and standard deviations across different attention heads. A significant bias vector exists in the embedding space.} 
\label{fig:rethinking-2}
\vspace{-3mm}
\end{figure}

\noindent\textbf{Concentrated distribution of embedding vectors.} For each attention layer of the transformer, we extract the query and key embeddings with shape $Q, K \in \mathbb{R}^{B \times S \times H \times D}$. For analysis purposes, we fix the batch size to $B=1$ and partition the sequence dimension $S$ into six semantically meaningful components: $Q_\text{text}$, $Q_\text{edit}$, $Q_\text{source}$, and similarly $K_\text{text}$, $K_\text{edit}$, $K_\text{source}$. Here, $Q_\text{text}, K_\text{text} \in \mathbb{R}^{N_{\mathrm{text}} \times H \times D}$ and the remaining components belong to $\mathbb{R}^{N_{\mathrm{img}} \times H \times D}$. 
We apply $L2$ normalization along the $N_{\mathrm{text}}$ or $N_{\mathrm{img}}$ dimension, reducing each component to a representation in $E \in \mathbb{R}^{H \times D}$, where each element $E_{h,d}$ represents the norm of 
the corresponding component in head $h$ and dimension $d$. 
Taking $Q_\mathrm{edit}$ as an example, $E_{h,d}$ is computed as:

\begin{equation}
E_{h,d} = \|Q_{:,h,d}\|_{2} 
= \sqrt{\sum_{s=1}^{N_{\mathrm{img}}} Q_{s,h,d}^{2}}
\end{equation}
The visualization results of $E$ are shown in Figure~\ref{fig:rethinking-1}. In the embedding vector space, each dimension index corresponds to a component, where the dark red regions in Figure~\ref{fig:rethinking-1} indicate positions with larger magnitudes that contribute more to the inner product between different token embeddings. By examining the relationship between RoPE~(Rotary Position Embedding~\cite{su2024roformer}) and dimension indices, we observe that text embeddings concentrate in low-frequency components associated with semantics, while image embeddings concentrate in high-frequency components capturing spatial relations. This finding suggests that the two modalities are not fully aligned in the shared embedding space.
Furthermore, we investigate the distribution of token embeddings in the vector space. Figure~\ref{fig:rethinking-2} presents the mean vector magnitudes and standard deviations across different attention heads, further revealing the presence of a significant bias vector among tokens in the embedding space.

\begin{figure*}[!t]
\centering
\includegraphics[width=0.95\linewidth]{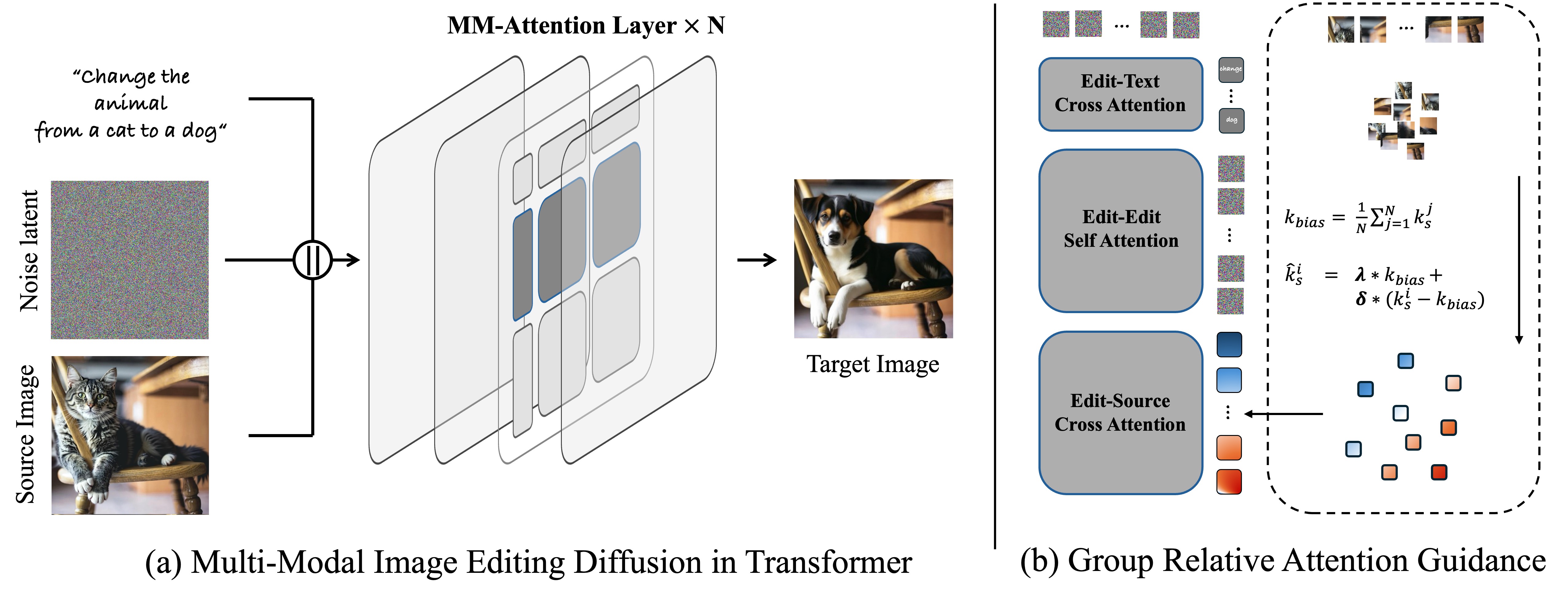}
\vspace{-3mm}
\caption{An illustration of applying Group Relative Attention Guidance in the MM-DiT image editing model. (a) The MM-Attention map corresponding to the query $Q_\mathrm{e}$, where GRAG is applied. (b) The processing of relative modulation to the source image's key embeddings. Red denotes enhanced tokens, while blue denotes suppressed tokens.} 
\vspace{-3mm}
\label{fig:method}
\end{figure*}

\noindent\textbf{Analysis of the bias vector.} The above findings suggest that the query and key embeddings in the attention layer exhibit a decomposable structure, where each can be represented as the sum of a dominant bias component and an independent variation:
\begin{equation}
q_i = q_{\text{bias}} + \Delta q_i, \quad k_i = k_{\text{b}} + \Delta k_i
\end{equation}
We also observe that the feature distributions of the same layer remain highly similar across different time steps and input samples. Based on this phenomenon, we hypothesize that the bias vector $q_{\text{bias}}, k_{\text{bias}}$ is related to the model weights and represents a fixed “\textit{editing action}” during the image editing process, while the variations of individual tokens relative to this bias vector correspond to the “\textit{content}” being edited.
Based on Equation~\ref{eq:unified-softmax}, we can derive:
\begin{equation}
\begin{aligned}
&S_{\mathrm{edit}}^{(i,\,j)} =\frac{e^{\langle q_e^i,k_t^{bias} \rangle } e^{\langle q_e^i,\Delta k_{t}^j \rangle }}{
e^{\langle q_e^i,k_t^{bias} \rangle }\Sigma_{t} + 
e^{\langle q_e^i,k_e^{bias} \rangle }\Sigma_{e} + 
e^{\langle q_e^i,k_s^{bias} \rangle }\Sigma_{s}},
\end{aligned}
\label{eq:unified-softmax-bias}
\end{equation}
\noindent\textit{Note:} For simplicity, the $\Sigma_{t}=\textstyle\sum_{p=1}^{N_{\mathrm{txt}}} e^{{\langle q_e^i,\Delta k_t^p \rangle }}, \Sigma_{e}=\textstyle\sum_{p=1}^{N_{\mathrm{img}}} e^{{\langle q_e^i,\Delta k_e^p \rangle }}, \Sigma_{s}=\textstyle\sum_{p=1}^{N_{\mathrm{img}}} e^{{\langle q_e^i,\Delta k_s^p \rangle }}$.

A strong shared bias component in both query and key embedding can dilute the influence of $\Delta k$, thereby reducing the sensitivity of attention scores to specific semantic differences. This insight naturally suggests that by modulating the magnitude of $\Delta k$, one can effectively control the extent to which the conditioning signals (e.g., edit instructions) influence the final output.

\section{Group Relative Attention Guidance}
\label{sec:GRAG}

The variations between individual token embeddings and the bias vector reflect how the editing content relates to the current layer's \textit{editing action}. By modulating their relative relationship, it becomes possible to achieve accurate and continuous control over the editing instructions. Based on this insight, we propose Group Relative Attention Guidance (GRAG). As illustrated in Figure~\ref{fig:method}, we modify the cross-attention component of the MM-Attention corresponding to the query $Q_\mathrm{e}$. In Figure~\ref{fig:method}, $K_\mathrm{s}$ is selected as a group of tokens, to which group-relative modulation is applied.

\begin{algorithm}[h]
\caption{Group-Relative Attention Guidance}
\label{alg:grag}
\begin{algorithmic}[1]
\Require Embedding $Q,K,V \in \mathbb{R}^{B \times S \times H \times D}$, token index $i_{start}, i_{end}$, guidance scale $\lambda, \delta$.
\Ensure Updated attention $\hat{A}$.
\State $Q,K,V = RoPE(Q), RoPE(K), V$ 
\State \textcolor{red}{$K_{s} \gets K[:,i_{start}:i_{end},:,:]$}
\State \textcolor{red}{$K_{bias} \gets mean(K_{s},dim=1)$}
\State \textcolor{red}{$K_{\Delta} \gets K_{s} - K_{bias}$}
\State \textcolor{red}{$K[:,i_{start}:i_{end},:,:] \gets \lambda * K_{bias} + \delta * K_{\Delta}$}
\State $\hat{A} \gets Attention(Q,K,V)$

\end{algorithmic}
\end{algorithm}

Formally, let $k_\mathrm{s}^i$ denote the conditional key embedding corresponding to token $i$, where $i=1,\dots,N_\mathrm{img}$. We first compute a group-level bias component as the mean of all conditional keys:
\begin{equation}
k_{\text{bias}} = \frac{1}{N_\mathrm{img}} \sum_{j=1}^{N_\mathrm{img}} k_\mathrm{s}^j
\end{equation}
The deviation of each token from this bias is then defined as:
\begin{equation}
\Delta k^i = k_\mathrm{s}^i - k_{\text{bias}}
\end{equation}
To control the influence of token-level variations, we introduce a tunable parameter $\lambda$ that scales these deviations:
\begin{equation}
\hat{k}_\mathrm{s}^i = \lambda \cdot k_{\text{bias}} + \delta \cdot \left(k_\mathrm{s}^i - k_{\text{bias}}\right)
\label{eq:group_relative}
\end{equation}
where $\hat{k}_\mathrm{s}^i$ denotes the updated key embedding under group relative attention guidance.

\begin{figure*}[t]
\centering
\includegraphics[width=0.99\linewidth]{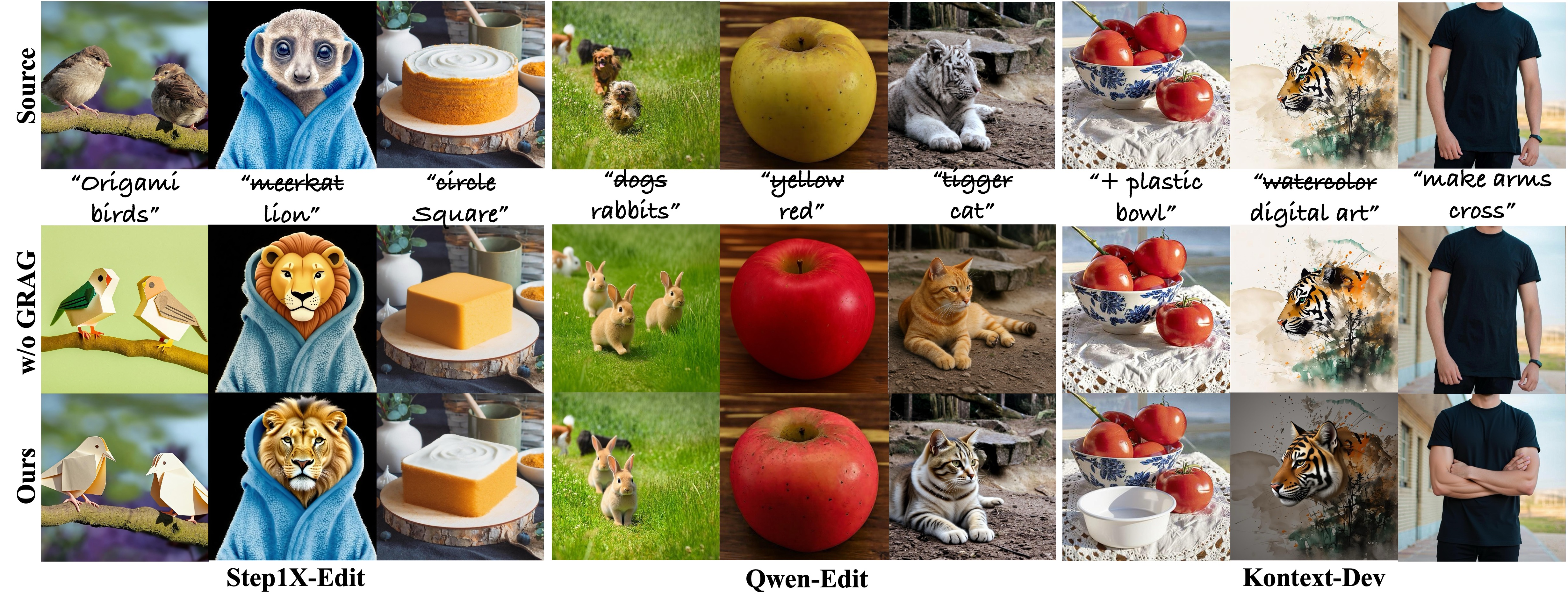}
\vspace{-5mm}
\caption{Visualization results on training-based image editing method. } 
\label{fig:comparison-1}
\vspace{-3mm}
\end{figure*}

\begin{figure*}[h]
\centering
\includegraphics[width=0.99\linewidth]{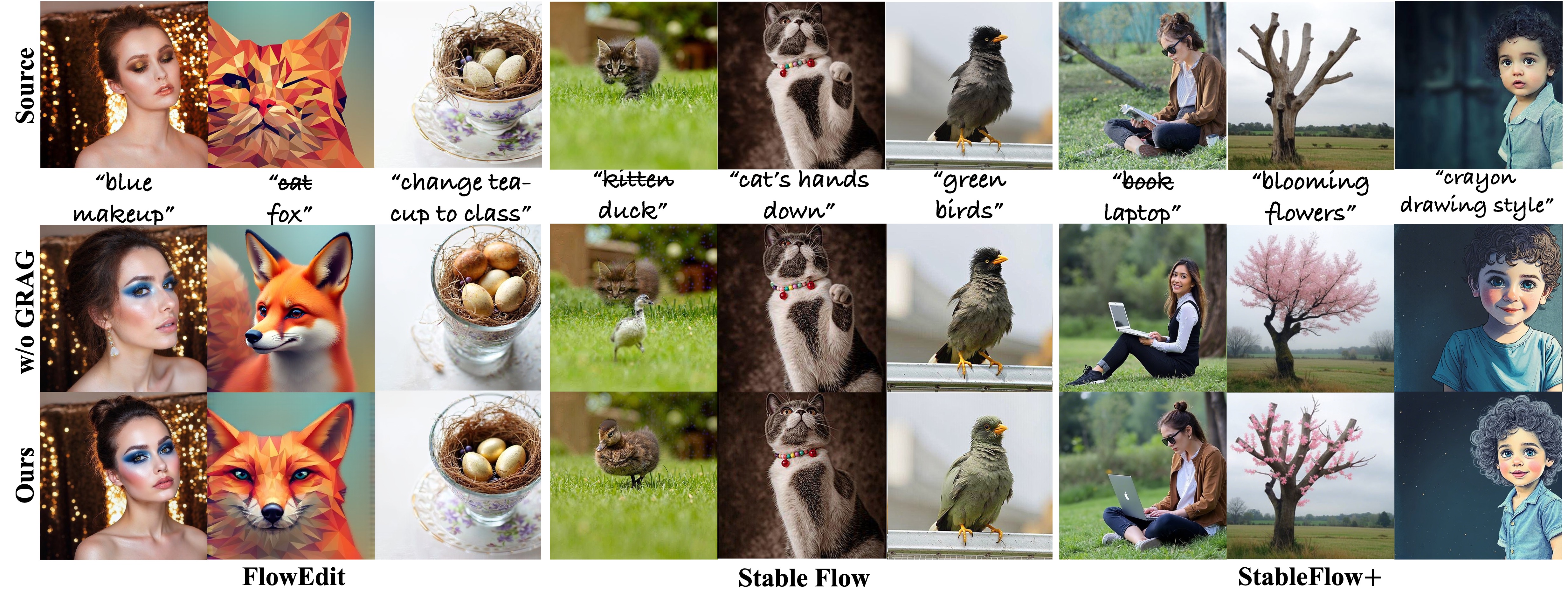}
\vspace{-5mm}
\caption{Visualization results on training-free image editing method. We update the original first-order inversion in StableFlow with a second-order ODE inversion method\cite{rfsolver,rf-inversion}, referred to as StableFlow$+$.} 
\label{fig:comparison-1-free}
\vspace{-5mm}
\end{figure*}

The scaling factor, $\lambda$ and $\delta$, are introduced to modulate the balance between the shared bias and token-specific variations. Both $\lambda$ and $\delta$ are positive real numbers. Specifically, $\lambda > 1$ enhances the influence of the selected tokens on the final image content, while $\lambda < 1$ reduces their impact. On the other hand, $\delta$ adjusts the focus intensity towards the selected tokens: $\delta > 1$ results in a more concentrated and precise editing impact, whereas $\delta < 1$ leads to a more diffused editing effect. The pseudo-code of Group Relative Attention Guidance is presented in Algorithm~\ref{alg:grag}, which consists of only four lines and can be seamlessly integrated into existing methods.


\section{Experiment}
\vspace{-1mm}
\subsection{Experiment Setting}
\vspace{-1mm}
\noindent\textbf{Baselines.} We apply the fixed GRAG parameters across six image editing models to validate GRAG's modulation effectiveness.
Kontext~\cite{labs2025flux}, Step1X-Edit~\cite{liu2025step1x} and Qwen-Edit~\cite{wu2025qwen} are Text-Image to Image (TI2I) editing method. And three training-free Text to Image (T2I) editing methods based on Flux.1-Dev\cite{blackforestlabs2024flux1dev} models (Flowedit~\cite{Flowedit}, Stableflow~\cite{stableflow}, Stableflow$+$). In these methods, GRAG is applied to the attention layers where source image features are injected. For training-based methods, GRAG is applied to all timesteps and layers, while for training-free methods, it is applied only during attention injection.
The random seed is fixed to 42. All experiments are conducted with a batch size of 1 and 24 inference steps. 

We compare our approach with the mainstream guidance method: Classifier-Free Guidance (CFG) and two attention control methods. Gated Attention~\cite{freeflux} (denoted as Attn-gate) controls the editing strength by setting a threshold on the attention correlation values computed between tokens, thereby limiting the number of tokens that respond to the editing instruction. In contrast, Attention Reweighting~\cite{saas} (Attn-weight) adjusts the editing strength by reweighting the attention scores produced by the SoftMax layer. All controllable editing strength analysis experiments are conducted based on the Qwen-Edit model.

\noindent\textbf{Evaluation.} We evaluate our method on PIE~\cite{pnp}. This benchmark covers a diverse range of editing tasks, including object addition/removal, style transfer, and pose modification. For quantitative evaluation, we adopt two complementary perspectives. Following previous works, we adopt LPIPS\cite{LPIPS} and SSIM\cite{SSIM} as quantitative metrics to evaluate the content preservation ability in non-edited regions. To assess the alignment between editing results and human preference, we employ the image editing reward model EditScore\cite{EditScore}. EditScore measures three aspects: consistency with the original image (Cons), prompt following (PF), and overall edit score (EditScore).

\begin{figure*}[t]
\centering
\includegraphics[page=1,width=\linewidth]{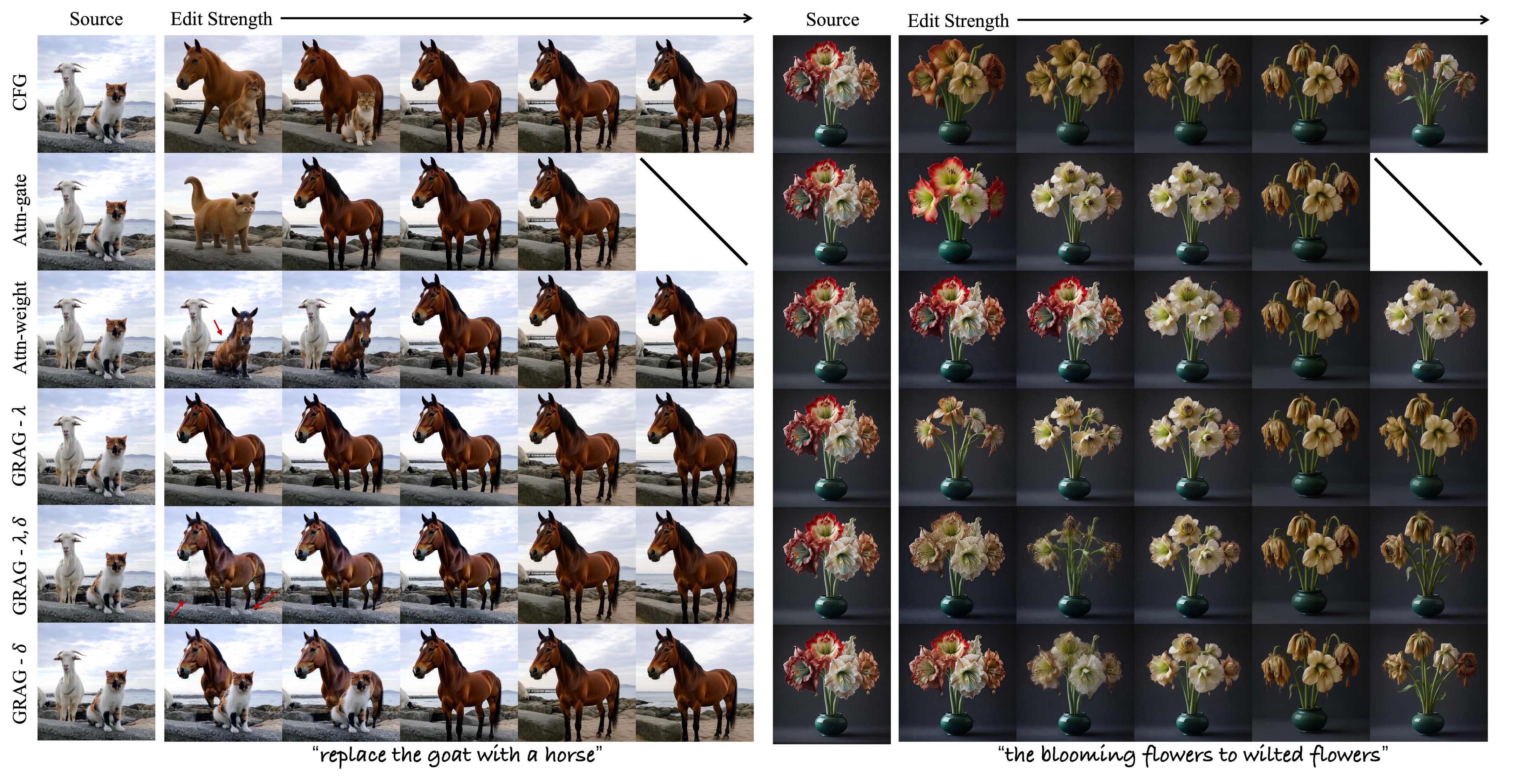}
\vskip -0.1in
\caption{Visualization results of different editing strength control methods. Attn-gate controls edits by limiting token activation and cannot enhance existing results (replaced with blanks). Red arrows mark visual artifacts. Overall, adjusting $\delta$ achieves the most continuous and precise editing control.} 
\label{fig:comparison-2}
\vspace{-3mm}
\end{figure*}

\subsection{Qualitative Analysis}
We apply GRAG to three mainstream MM-DiT-based image editing models, with qualitative results shown in Figure~\ref{fig:comparison-1}. On Step1X-Edit and Qwen-Edit, our method improves consistency between the edited images and the original references while preserving the intended editing effects, yielding more realistic and natural outcomes. Since Step1X-Edit and Qwen-Edit leverage vision–language models to encode editing instructions, the additional instruction information often enhances responsiveness but reduces consistency. We select the source image tokens as group and apply GRAG to enhance the response of edit-related tokens to the editing instructions while suppressing the response of irrelevant tokens.
For instance, in the first column of Figure~\ref{fig:comparison-1}, GRAG successfully changes the texture of the bird while retaining the details of the tree trunk; in the fifth column, it alters the color of the apple while preserving fine-grained surface details. These examples demonstrate the ability of GRAG to achieve precise and continuous control over edits while maintaining fidelity to the source image. For the original Kontext model, we select the text tokens as the group and apply GRAG to enhance the model’s response to the editing instructions. As shown on the right side of Figure~\ref{fig:comparison-1}, the baseline fails to respond to the editing instruction, with no change in content, whereas applying GRAG enables successful editing.

As shown in Figure~\ref{fig:comparison-1-free}, our approach achieves adjustment of the editing results, indicating that GRAG remains effective in training-free editing method.
\begin{figure*}[h]
\vspace{-1mm}
\centering
\includegraphics[width=1.0\linewidth]{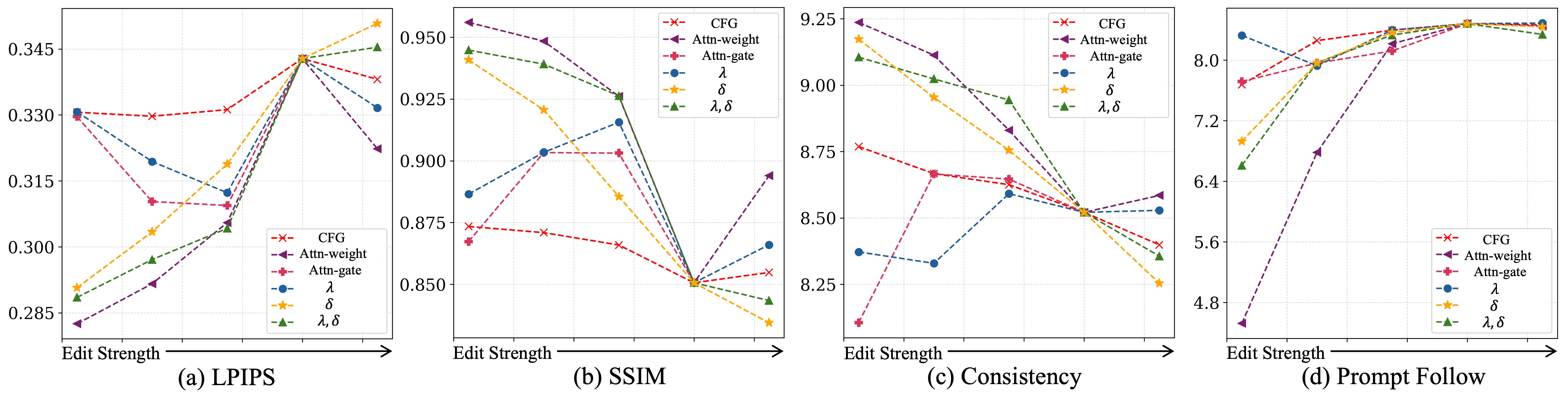}
\vspace{-5mm}
\caption{Comparison of different guidance strategies under varying guidance strengths. The data in the line chart correspond to Table~\ref{tab:abla}. The $\delta$ parameter yields the most continuous and effective editing guidance.} 
\label{fig:ablation-vis}
\vspace{0mm}
\end{figure*}

\begin{table}[h]

    \centering
    \resizebox{\linewidth}{!}{
    \begin{tabular}{l|cccc:c}
    \toprule
    Model & LPIPS$\downarrow$ & SSIM$\uparrow$ & \text{Cons}$\uparrow$ & PF$\uparrow$ &\text{EditScore}$\uparrow$ \\
    \hline
    \textbf{Training-Based} & & &&  & \\
    \hline
    Kontext-Dev & 0.3061 & 0.9213 & 8.9051 & 6.9051 & 6.0887 \\
    +GRAG       & 0.3873 & 0.8156 & 8.6788 & 7.4177 & \textbf{6.4081} \\
    \hline
    Step1X-Edit & 0.3228 & 0.9042 & 8.4714 & 7.8406 & 6.8292 \\
    +GRAG       & 0.3174 & 0.9137 & 8.6240 & 8.0406 & \textbf{7.0045} \\
    \hline
    Qwen-Edit   & 0.3428 & 0.8506 & 8.5211 & 8.4806 & 7.2576 \\
    +GRAG       & 0.3042 & 0.9263 & 8.9440 & 8.3303 & \textbf{7.3245} \\
    \hline
    \textbf{Training-Free} & & &&  & \\
    \hline
    Flowedit & 0.3758 & 0.8237 & 6.8794 & 5.0531 & 4.6635 \\
    +GRAG     & 0.3670 & 0.8312 & 7.2223 & 4.8954 & \textbf{4.6697} \\
    \hline
    StableFlow & 0.3219 & 0.9185 & 8.9309 & 2.2177 & 2.4573 \\
    +GRAG       & 0.3292 & 0.9098 & 8.8731 & 2.7429 & \textbf{3.0303} \\
    \hline
    StableFlow$+$  & 0.3691 & 0.8229 & 7.3599 & 5.3926 & 5.0970 \\
    +GRAG       & 0.3595 & 0.8316 & 7.7997 & 4.8395 & 4.7251 \\
    
    \bottomrule
    \end{tabular}
    }
    \vspace{-2mm}
    \caption{Quantitative results on different image editing methods.}
    \label{tab:comparsion-pie}
    \vspace{-5mm}
\end{table}

\subsection{Quantitative Analysis}
As shown in Tab~\ref{tab:comparsion-pie}, we perform quantitative evaluations on the PIE dataset. Step1X-Edit and Qwen-Edit exhibit enhanced consistency between the edited outputs and the original images after integrating GRAG, as indicated by improvements in LPIPS, SSIM, and Cons. Although a slight decline is observed in PF, the overall EditScore, which reflects overall editing quality, increases. In contrast, Kontext demonstrates a noticeable improvement in PF and achieves a higher EditScore after applying GRAG. These trends align well with the visual results.

As shown in Table~\ref{tab:comparsion-pie}-Bottom, GRAG also provides performance gains on training-free editing methods. However, its adaptability is relatively lower compared to training-based methods.
We attribute this to the fact that GRAG primarily modulates the cross-attention component in MM-Attention (Figure~\ref{fig:method}), whereas in untrained T2I models, source image features are introduced through the edit–edit self-attention branch (Figure~\ref{fig:method}-b). We provide further discussion in the Supplementary Material.

\begin{table}[h]
    \centering

  \resizebox{1\linewidth}{!}{
  \begin{tabular}{ c | cccc:c}
  \toprule
  Method          & LPIPS $\downarrow$  & SSIM$\uparrow$  & Cons $\uparrow$  & PF $\uparrow$ & EditScore $\uparrow$\\ \hline
  
  CFG = 5.00 & 0.3381 & 0.8548 & 8.3989 & 8.4640 & 7.1857\\
  \rowcolor{gray!25}CFG = 4.00 & 0.3428 & 0.8506 & 8.5211 & 8.4806 & 7.2576\\
  CFG = 3.00 & 0.3312 & 0.8659 & 8.6251 & 8.3954 & 7.2761\\
  CFG = 2.00 & 0.3297 & 0.8709 & 8.6669 & 8.2566 & 7.2247\\
  CFG = 1.00 & 0.3306 & 0.8734 & 8.7686 & 7.6760 & 6.8294\\ \hline

  Attn-weight = 0.8 & 0.3223 & 0.8940 & 8.5846 & 8.4503 & 7.2808\\
  \rowcolor{gray!25}Attn-weight = 1.0 & 0.3428 & 0.8506 & 8.5211 & 8.4806 & 7.2576\\
  Attn-weight = 7.0 & 0.3055 & 0.9261 & 8.8301 & 8.2159 & 7.2214\\
  Attn-weight = 10.0 & 0.2916 & 0.9484 & 9.1130 & 6.7797 & 5.9555\\
  Attn-weight = 16.0 & 0.2825 & 0.9560 & 9.2361 & 4.5230 & 3.9953\\ \hline

  \rowcolor{gray!25}Attn-gate = 1.0 & 0.3428 & 0.8506 & 8.5211 & 8.4806 & 7.2576\\
  Attn-gate = 0.9 & 0.3094 & 0.9031 & 8.6460 & 8.1206 & 7.4004\\
  Attn-gate = 0.7 & 0.3103 & 0.9033 & 8.6655 & 7.9611 & 7.2531\\
  Attn-gate = 0.3 & 0.3296 & 0.8673 & 8.1051 & 7.7184 & 4.6505\\ \hline
  
  $\lambda=0.95,\delta=1.00$ & 0.3316 & 0.8660 & 8.5286 & 8.4886 & 7.2725\\
  \rowcolor{gray!25}$\lambda=1.00,\delta=1.00$ & 0.3428 & 0.8506 & 8.5211 & 8.4806 & 7.2576\\
  $\lambda=1.05,\delta=1.00$ & 0.3123 & 0.9156 & 8.5914 & 8.3977 & 7.2990\\
  $\lambda=1.10,\delta=1.00$ & 0.3194 & 0.9034 & 8.3291 & 7.9251 & 7.1992\\
  $\lambda=1.15,\delta=1.00$ & 0.3307 & 0.8865 & 8.3720 & 8.3269 & 7.1863\\ \hline
  
  $\lambda=1.00,\delta=0.95$ & 0.3508 & 0.8344 & 8.2543 & 8.4394 & 7.1991\\
  \rowcolor{gray!25}$\lambda=1.00,\delta=1.00$ & 0.3428 & 0.8506 & 8.5211 & 8.4806 & 7.2576\\
  $\lambda=1.00,\delta=1.05$ & 0.3188 & 0.8855 & 8.7549 & 8.3651 & 7.2679\\
  $\lambda=1.00,\delta=1.10$ & 0.3034 & 0.9206 & 8.9537 & 7.9611 & 6.9872\\
  $\lambda=1.00,\delta=1.15$ & 0.2907 & 0.9408 & 9.1731 & 6.9291 & 6.1730\\ \hline
  
  $\lambda=0.95,\delta=0.95$ & 0.3454 & 0.8434 & 8.3560 & 8.3394 & 7.2162\\
  \rowcolor{gray!25}$\lambda=1.00,\delta=1.00$ & 0.3428 & 0.8506 & 8.5211 & 8.4806 & 7.2576\\
  $\lambda=1.05,\delta=1.05$ & 0.3042 & 0.9263 & 8.9440 & 8.3303 & 7.3245\\
  $\lambda=1.10,\delta=1.10$ & 0.2971 & 0.9391 & 9.0234 & 7.9674 & 7.0243\\
  $\lambda=1.15,\delta=1.15$ & 0.2885 & 0.9448 & 9.1051 & 6.6091 & 5.9955\\
  \bottomrule
  \end{tabular}}
  \vspace{-2mm}
  \caption{Continuity and effectiveness analysis of different editing strength control methods. Appropriate parameters are selected for each method within its effective working range.
}
  \vspace{-5mm}
  \label{tab:abla}
  \end{table}

\subsection{Ablation Study}

\noindent\textbf{Comparison with other guidance method.} 
Although Classifier-Free Guidance (CFG) can influence the generation results by adjusting the denoising process, it provides only limited control over editing strength and often leads to a noticeable degradation in image quality when the guidance scale is small (e.g., CFG = 1). Both attention-based control methods produce visible effects on the edited results; however, Attn-gate exhibits discontinuous control and tends to generate distorted outputs, as shown in the second row on the left of Figure~\ref{fig:comparison-2}. Attn-weight achieves a modulation effect most similar to ours, but it may cause incorrect edits—for example, the cat is mistakenly replaced by a horse in Figure~\ref{fig:comparison-2}—and it still fails to ensure smooth and consistent editing, where the flowers first wither and then become unnaturally vibrant as the editing strength increases.
As shown in Table~\ref{tab:abla} and Figure~\ref{fig:ablation-vis}, GRAG enables precise and continuous control over the editing, producing smooth and consistent adjustments as the editing strength increases, the visual comparsion shown in Figure~\ref{fig:comparison-2}. Such controllability is crucial for customized image editing applications.

\noindent\textbf{Effectiveness Analysis of Group Relative.} We analyze the influence of the $\lambda$ and $\delta$ parameters in Eq.~\ref{eq:group_relative} on the editing results. 
Three groups of experiments are conducted: only $\lambda$, only $\delta$, and both $\lambda$ and $\delta$ simultaneously. 
The qualitative results are shown in Figure~\ref{fig:comparison-2}, while the quantitative results on the PIE benchmark are presented in Table~\ref{tab:abla} and Figure~\ref{fig:ablation-vis}. 
Adjusting $\lambda$ alone shows no significant impact on the editing results, corresponding to the fluctuating curves in Figure~\ref{fig:ablation-vis}, which indicates that tuning $\lambda$ cannot effectively control the editing strength. 
In contrast, jointly adjusting $\lambda$ and $\delta$ enables a certain degree of controllable editing but fails to achieve continuous precision. Moreover, this simultaneous adjustment often degrades visual fidelity, leading to undesirable artifacts such as the distorted flowers and the visible artifacts of the horse sample in Figure~\ref{fig:comparison-2}. 
Adjusting $\delta$ alone yields the best results, corresponding to the smoothest metric variation in Figure~\ref{fig:ablation-vis} and the most continuous editing transitions in Figure~\ref{fig:comparison-2}.

\vspace{-3mm}
\section{Conclusion}
In this work, we revisited the internal attention mechanism of Diffusion-in-Transformer (DiT) models and revealed the presence of a shared bias vector that governs editing behavior. Building on this insight, we introduced Group Relative Attention Guidance (GRAG), a lightweight yet effective strategy that modulates token deviations from the group bias to achieve fine-grained and continuous control over editing strength. GRAG can be seamlessly integrated into existing DiT-based editors, consistently improving both controllability and fidelity. Our findings provide new insights into the internal dynamics of multi-modal attention and offer a practical direction for enhancing controllable image editing in future DiT architectures.

{
    \small
    \bibliographystyle{ieeenat_fullname}
    \bibliography{main}

@String(CVPR= {IEEE Conf. Comput. Vis. Pattern Recog.})

@String(ECCV= {Eur. Conf. Comput. Vis.})

@String(ICLR = {Int. Conf. Learn. Represent.})

@String(CVPR  = {CVPR})

@String(ECCV  = {ECCV})

@String(ICLR  = {ICLR})

@article{gal2022image,
  title={An image is worth one word: Personalizing text-to-image generation using textual inversion},
  author={Gal, Rinon and Alaluf, Yuval and Atzmon, Yuval and Patashnik, Or and Bermano, Amit H and Chechik, Gal and Cohen-Or, Daniel},
  journal={arXiv preprint arXiv:2208.01618},
  year={2022}
}

@article{hu2022lora,
  title={Lora: Low-rank adaptation of large language models.},
  author={Hu, Edward J and Shen, Yelong and Wallis, Phillip and Allen-Zhu, Zeyuan and Li, Yuanzhi and Wang, Shean and Wang, Lu and Chen, Weizhu and others},
  journal={ICLR},
  volume={1},
  number={2},
  pages={3},
  year={2022}
}

@misc{gemini,
  author       = {Gemini Team and others},
  title        = {Gemini: A Family of Highly Capable Multimodal Models},
  year         = {2023},
  howpublished = {arXiv preprint arXiv:2312.11805},
  url          = {https://arxiv.org/abs/2312.11805}
}

@misc{midjourney,
  author       = {{Midjourney}},
  title        = {Midjourney},
  year         = {2022},
  howpublished = {\url{https://www.midjourney.com}},
}

@misc{runway,
  author       = {{Runway}},
  title        = {Runway},
  year         = {2023},
  howpublished = {\url{https://runwayml.com}},
}

@misc{gpt4v,
  author       = {OpenAI},
  title        = {GPT-4V(ision) System Card},
  year         = {2023},
  howpublished = {\url{https://openai.com/research/gpt-4v-system-card}},
}

@article{zhang2025context,
  title={In-context edit: Enabling instructional image editing with in-context generation in large scale diffusion transformer},
  author={Zhang, Zechuan and Xie, Ji and Lu, Yu and Yang, Zongxin and Yang, Yi},
  journal={arXiv preprint arXiv:2504.20690},
  year={2025}
}

@article{cai2025hidream,
  title={HiDream-I1: A High-Efficient Image Generative Foundation Model with Sparse Diffusion Transformer},
  author={Cai, Qi and Chen, Jingwen and Chen, Yang and Li, Yehao and Long, Fuchen and Pan, Yingwei and Qiu, Zhaofan and Zhang, Yiheng and Gao, Fengbin and Xu, Peihan and others},
  journal={arXiv preprint arXiv:2505.22705},
  year={2025}
}

@inproceedings{xiao2025omnigen,
  title={Omnigen: Unified image generation},
  author={Xiao, Shitao and Wang, Yueze and Zhou, Junjie and Yuan, Huaying and Xing, Xingrun and Yan, Ruiran and Li, Chaofan and Wang, Shuting and Huang, Tiejun and Liu, Zheng},
  booktitle={Proceedings of the Computer Vision and Pattern Recognition Conference},
  pages={13294--13304},
  year={2025}
}

@inproceedings{sheynin2024emu,
  title={Emu edit: Precise image editing via recognition and generation tasks},
  author={Sheynin, Shelly and Polyak, Adam and Singer, Uriel and Kirstain, Yuval and Zohar, Amit and Ashual, Oron and Parikh, Devi and Taigman, Yaniv},
  booktitle={Proceedings of the IEEE/CVF Conference on Computer Vision and Pattern Recognition},
  pages={8871--8879},
  year={2024}
}

@article{su2024roformer,
  title={Roformer: Enhanced transformer with rotary position embedding},
  author={Su, Jianlin and Ahmed, Murtadha and Lu, Yu and Pan, Shengfeng and Bo, Wen and Liu, Yunfeng},
  journal={Neurocomputing},
  volume={568},
  pages={127063},
  year={2024},
  publisher={Elsevier}
}

@article{jin2025massive,
  title={Massive Values in Self-Attention Modules are the Key to Contextual Knowledge Understanding},
  author={Jin, Mingyu and Mei, Kai and Xu, Wujiang and Sun, Mingjie and Tang, Ruixiang and Du, Mengnan and Liu, Zirui and Zhang, Yongfeng},
  journal={arXiv preprint arXiv:2502.01563},
  year={2025}
}

@inproceedings{ruiz2023dreambooth,
  title={Dreambooth: Fine tuning text-to-image diffusion models for subject-driven generation},
  author={Ruiz, Nataniel and Li, Yuanzhen and Jampani, Varun and Pritch, Yael and Rubinstein, Michael and Aberman, Kfir},
  booktitle={Proceedings of the IEEE/CVF conference on computer vision and pattern recognition},
  pages={22500--22510},
  year={2023}
}

@article{labs2025flux,
  title={FLUX. 1 Kontext: Flow Matching for In-Context Image Generation and Editing in Latent Space},
  author={Labs, Black Forest and Batifol, Stephen and Blattmann, Andreas and Boesel, Frederic and Consul, Saksham and Diagne, Cyril and Dockhorn, Tim and English, Jack and English, Zion and Esser, Patrick and others},
  journal={arXiv preprint arXiv:2506.15742},
  year={2025}
}

@article{chen2023pixart,
  title={Pixart-$\alpha$: Fast training of diffusion transformer for photorealistic text-to-image synthesis},
  author={Chen, Junsong and Yu, Jincheng and Ge, Chongjian and Yao, Lewei and Xie, Enze and Wu, Yue and Wang, Zhongdao and Kwok, James and Luo, Ping and Lu, Huchuan and others},
  journal={arXiv preprint arXiv:2310.00426},
  year={2023}
}

@article{wu2025qwen,
  title={Qwen-image technical report},
  author={Wu, Chenfei and Li, Jiahao and Zhou, Jingren and Lin, Junyang and Gao, Kaiyuan and Yan, Kun and Yin, Sheng-ming and Bai, Shuai and Xu, Xiao and Chen, Yilei and others},
  journal={arXiv preprint arXiv:2508.02324},
  year={2025}
}

@article{shao2024deepseekmath,
  title={Deepseekmath: Pushing the limits of mathematical reasoning in open language models},
  author={Shao, Zhihong and Wang, Peiyi and Zhu, Qihao and Xu, Runxin and Song, Junxiao and Bi, Xiao and Zhang, Haowei and Zhang, Mingchuan and Li, YK and Wu, Yang and others},
  journal={arXiv preprint arXiv:2402.03300},
  year={2024}
}

@article{TACA,
  title={Rethinking Cross-Modal Interaction in Multimodal Diffusion Transformers},
  author={Lv, Zhengyao and Pan, Tianlin and Si, Chenyang and Chen, Zhaoxi and Zuo, Wangmeng and Liu, Ziwei and Wong, Kwan-Yee K},
  journal={arXiv preprint arXiv:2506.07986},
  year={2025}
}

@article{ddpm,
  title={Denoising diffusion probabilistic models},
  author={Ho, Jonathan and Jain, Ajay and Abbeel, Pieter},
  journal={Advances in neural information processing systems},
  volume={33},
  pages={6840--6851},
  year={2020}
}

@inproceedings{ldm,
  title={High-resolution image synthesis with latent diffusion models},
  author={Rombach, Robin and Blattmann, Andreas and Lorenz, Dominik and Esser, Patrick and Ommer, Bj{\"o}rn},
  booktitle={Proceedings of the IEEE/CVF conference on computer vision and pattern recognition},
  pages={10684--10695},
  year={2022}
}

@article{podell2023sdxl,
  title={Sdxl: Improving latent diffusion models for high-resolution image synthesis},
  author={Podell, Dustin and English, Zion and Lacey, Kyle and Blattmann, Andreas and Dockhorn, Tim and M{\"u}ller, Jonas and Penna, Joe and Rombach, Robin},
  journal={arXiv preprint arXiv:2307.01952},
  year={2023}
}

@inproceedings{sd3,
  title={Scaling rectified flow transformers for high-resolution image synthesis},
  author={Esser, Patrick and Kulal, Sumith and Blattmann, Andreas and Entezari, Rahim and M{\"u}ller, Jonas and Saini, Harry and Levi, Yam and Lorenz, Dominik and Sauer, Axel and Boesel, Frederic and others},
  booktitle={Forty-first international conference on machine learning},
  year={2024}
}

@misc{blackforestlabs2024flux1dev,
  title        = {FLUX.1-dev},
  author       = {{Black Forest Labs}},
  year         = {2024},
  howpublished = {\url{https://huggingface.co/black-forest-labs/FLUX.1-dev}}
}

@article{pnp,
  title={PnP Inversion: Boosting Diffusion-based Editing with 3 Lines of Code},
  author={Ju, Xuan and Zeng, Ailing and Bian, Yuxuan and Liu, Shaoteng and Xu, Qiang},
  journal={International Conference on Learning Representations ({ICLR})},
  year={2024}
}

@article{rf-solver,
  title={Taming Rectified Flow for Inversion and Editing},
  author={Wang, Jiangshan and Pu, Junfu and Qi, Zhongang and Guo, Jiayi and Ma, Yue and Huang, Nisha and Chen, Yuxin and Li, Xiu and Shan, Ying},
  journal={arXiv preprint arXiv:2411.04746},
  year={2024}
}

@article{attention,
  title={Attention is all you need},
  author={Vaswani, Ashish and Shazeer, Noam and Parmar, Niki and Uszkoreit, Jakob and Jones, Llion and Gomez, Aidan N and Kaiser, {\L}ukasz and Polosukhin, Illia},
  journal={Advances in neural information processing systems},
  volume={30},
  year={2017}
}

@inproceedings{brooks2023instructpix2pix,
  title={Instructpix2pix: Learning to follow image editing instructions},
  author={Brooks, Tim and Holynski, Aleksander and Efros, Alexei A},
  booktitle={Proceedings of the IEEE/CVF conference on computer vision and pattern recognition},
  pages={18392--18402},
  year={2023}
}

@article{stableflow,
  title={Stable Flow: Vital Layers for Training-Free Image Editing},
  author={Avrahami, Omri and Patashnik, Or and Fried, Ohad and Nemchinov, Egor and Aberman, Kfir and Lischinski, Dani and Cohen-Or, Daniel},
  journal={arXiv preprint arXiv:2411.14430},
  year={2024}
}

@article{t5,
  title={Exploring the limits of transfer learning with a unified text-to-text transformer},
  author={Raffel, Colin and Shazeer, Noam and Roberts, Adam and Lee, Katherine and Narang, Sharan and Matena, Michael and Zhou, Yanqi and Li, Wei and Liu, Peter J},
  journal={Journal of machine learning research},
  volume={21},
  number={140},
  pages={1--67},
  year={2020}
}

@inproceedings{clip,
  title={Learning transferable visual models from natural language supervision},
  author={Radford, Alec and Kim, Jong Wook and Hallacy, Chris and Ramesh, Aditya and Goh, Gabriel and Agarwal, Sandhini and Sastry, Girish and Askell, Amanda and Mishkin, Pamela and Clark, Jack and others},
  booktitle={International conference on machine learning},
  pages={8748--8763},
  year={2021},
  organization={PmLR}
}

@inproceedings{dit,
  title={Scalable diffusion models with transformers},
  author={Peebles, William and Xie, Saining},
  booktitle={Proceedings of the IEEE/CVF international conference on computer vision},
  pages={4195--4205},
  year={2023}
}

@article{liu2025step1x,
  title={Step1x-edit: A practical framework for general image editing},
  author={Liu, Shiyu and Han, Yucheng and Xing, Peng and Yin, Fukun and Wang, Rui and Cheng, Wei and Liao, Jiaqi and Wang, Yingming and Fu, Honghao and Han, Chunrui and others},
  journal={arXiv preprint arXiv:2504.17761},
  year={2025}
}

@article{Flowedit,
  author       = {Vladimir Kulikov and
                  Matan Kleiner and
                  Inbar Huberman{-}Spiegelglas and
                  Tomer Michaeli},
  title        = {FlowEdit: Inversion-Free Text-Based Editing Using Pre-Trained Flow
                  Models},
  journal      = {CoRR},
  volume       = {abs/2412.08629},
  year         = {2024},
  url          = {https://doi.org/10.48550/arXiv.2412.08629},
  doi          = {10.48550/ARXIV.2412.08629},
  eprinttype    = {arXiv},
  eprint       = {2412.08629},
  bibsource    = {dblp computer science bibliography, https://dblp.org}
}

@article{rfsolver,
  author       = {Jiangshan Wang and
                  Junfu Pu and
                  Zhongang Qi and
                  Jiayi Guo and
                  Yue Ma and
                  Nisha Huang and
                  Yuxin Chen and
                  Xiu Li and
                  Ying Shan},
  title        = {Taming Rectified Flow for Inversion and Editing},
  journal      = {CoRR},
  volume       = {abs/2411.04746},
  year         = {2024},
  url          = {https://doi.org/10.48550/arXiv.2411.04746},
  doi          = {10.48550/ARXIV.2411.04746},
  eprinttype    = {arXiv},
  eprint       = {2411.04746},
  bibsource    = {dblp computer science bibliography, https://dblp.org}
}

@article{EditScore,
  author       = {Xin Luo and
                  Jiahao Wang and
                  Chenyuan Wu and
                  Shitao Xiao and
                  Xiyan Jiang and
                  Defu Lian and
                  Jiajun Zhang and
                  Dong Li and
                  Zheng Liu},
  title        = {EditScore: Unlocking Online {RL} for Image Editing via High-Fidelity
                  Reward Modeling},
  journal      = {CoRR},
  volume       = {abs/2509.23909},
  year         = {2025},
  url          = {https://doi.org/10.48550/arXiv.2509.23909},
  doi          = {10.48550/ARXIV.2509.23909},
  eprinttype    = {arXiv},
  eprint       = {2509.23909},
  bibsource    = {dblp computer science bibliography, https://dblp.org}
}

@inproceedings{LPIPS,
  title={The Unreasonable Effectiveness of Deep Features as a Perceptual Metric},
  author={Zhang, Richard and Isola, Phillip and Efros, Alexei A and Shechtman, Eli and Wang, Oliver},
  booktitle={CVPR},
  year={2018}
}

@article{SSIM,
  title={Image quality assessment: from error visibility to structural similarity},
  author={Wang, Zhou and Bovik, Alan C and Sheikh, Hamid R and Simoncelli, Eero P},
  journal={IEEE transactions on image processing},
  volume={13},
  number={4},
  pages={600--612},
  year={2004},
  publisher={IEEE}
}

@inproceedings{boow-vton,
  author       = {Xuanpu Zhang and
                  Dan Song and
                  Pengxin Zhan and
                  Tianyu Chang and
                  Jianhao Zeng and
                  Qingguo Chen and
                  Weihua Luo and
                  An{-}An Liu},
  title        = {BooW-VTON: Boosting In-the-Wild Virtual Try-On via Mask-Free Pseudo
                  Data Training},
  booktitle    = {{IEEE/CVF} Conference on Computer Vision and Pattern Recognition,
                  {CVPR} 2025, Nashville, TN, USA, June 11-15, 2025},
  pages        = {26399--26408},
  publisher    = {Computer Vision Foundation / {IEEE}},
  year         = {2025},
  url          = {https://openaccess.thecvf.com/content/CVPR2025/html/Zhang\_BooW-VTON\_Boosting\_In-the-Wild\_Virtual\_Try-On\_via\_Mask-Free\_Pseudo\_Data\_Training\_CVPR\_2025\_paper.html},
  doi          = {10.1109/CVPR52734.2025.02458},
  bibsource    = {dblp computer science bibliography, https://dblp.org}
}

@inproceedings{rf-inversion,
  author       = {Litu Rout and
                  Yujia Chen and
                  Nataniel Ruiz and
                  Constantine Caramanis and
                  Sanjay Shakkottai and
                  Wen{-}Sheng Chu},
  title        = {Semantic Image Inversion and Editing using Rectified Stochastic Differential
                  Equations},
  booktitle    = {The Thirteenth International Conference on Learning Representations,
                  {ICLR} 2025, Singapore, April 24-28, 2025},
  publisher    = {OpenReview.net},
  year         = {2025},
  url          = {https://openreview.net/forum?id=Hu0FSOSEyS},
  bibsource    = {dblp computer science bibliography, https://dblp.org}
}

@inproceedings{DDIM,
  author       = {Jiaming Song and
                  Chenlin Meng and
                  Stefano Ermon},
  title        = {Denoising Diffusion Implicit Models},
  booktitle    = {9th International Conference on Learning Representations, {ICLR} 2021,
                  Virtual Event, Austria, May 3-7, 2021},
  publisher    = {OpenReview.net},
  year         = {2021},
  url          = {https://openreview.net/forum?id=St1giarCHLP},
  bibsource    = {dblp computer science bibliography, https://dblp.org}
}

@inproceedings{video-editing-lora,
  title={Separate motion from appearance: Customizing motion via customizing text-to-video diffusion models},
  author={Liu, Huijie and Wang, Jingyun and Ma, Shuai and Hu, Jie and Wei, Xiaoming and Kang, Guoliang},
  booktitle={Proceedings of the 33rd ACM International Conference on Multimedia},
  pages={9227--9236},
  year={2025}
}

@article{video-editing,
  title={FreeInv: Free Lunch for Improving DDIM Inversion},
  author={Bao, Yuxiang and Liu, Huijie and Gao, Xun and Fu, Huan and Kang, Guoliang},
  journal={arXiv preprint arXiv:2503.23035},
  year={2025}
}

@article{image-editing,
  title={Omni-Dish: Photorealistic and Faithful Image Generation and Editing for Arbitrary Chinese Dishes},
  author={Liu, Huijie and Wang, Bingcan and Hu, Jie and Wei, Xiaoming and Kang, Guoliang},
  journal={arXiv preprint arXiv:2504.09948},
  year={2025}
}

@inproceedings{concept_slider,
  author       = {Rohit Gandikota and
                  Joanna Materzynska and
                  Tingrui Zhou and
                  Antonio Torralba and
                  David Bau},
  editor       = {Ales Leonardis and
                  Elisa Ricci and
                  Stefan Roth and
                  Olga Russakovsky and
                  Torsten Sattler and
                  G{\"{u}}l Varol},
  title        = {Concept Sliders: LoRA Adaptors for Precise Control in Diffusion Models},
  booktitle    = {Computer Vision - {ECCV} 2024 - 18th European Conference, Milan, Italy,
                  September 29-October 4, 2024, Proceedings, Part {XL}},
  series       = {Lecture Notes in Computer Science},
  volume       = {15098},
  pages        = {172--188},
  publisher    = {Springer},
  year         = {2024},
  url          = {https://doi.org/10.1007/978-3-031-73661-2\_10},
  doi          = {10.1007/978-3-031-73661-2\_10},
  bibsource    = {dblp computer science bibliography, https://dblp.org}
}

@article{saas,
  author       = {Chao Zhou and
                  Tianyi Wei and
                  Nenghai Yu},
  title        = {Scale Your Instructions: Enhance the Instruction-Following Fidelity
                  of Unified Image Generation Model by Self-Adaptive Attention Scaling},
  journal      = {CoRR},
  volume       = {abs/2507.16240},
  year         = {2025},
  url          = {https://doi.org/10.48550/arXiv.2507.16240},
  doi          = {10.48550/ARXIV.2507.16240},
  eprinttype    = {arXiv},
  eprint       = {2507.16240},
  bibsource    = {dblp computer science bibliography, https://dblp.org}
}

@inproceedings{attr_control,
  author       = {Stefan Andreas Baumann and
                  Felix Krause and
                  Michael Neumayr and
                  Nick Stracke and
                  Melvin Sevi and
                  Vincent Tao Hu and
                  Bj{\"{o}}rn Ommer},
  title        = {Continuous, Subject-Specific Attribute Control in {T2I} Models by
                  Identifying Semantic Directions},
  booktitle    = {{IEEE/CVF} Conference on Computer Vision and Pattern Recognition,
                  {CVPR} 2025, Nashville, TN, USA, June 11-15, 2025},
  pages        = {13231--13241},
  publisher    = {Computer Vision Foundation / {IEEE}},
  year         = {2025},
  url          = {https://openaccess.thecvf.com/content/CVPR2025/html/Baumann\_Continuous\_Subject-Specific\_Attribute\_Control\_in\_T2I\_Models\_by\_Identifying\_Semantic\_CVPR\_2025\_paper.html},
  doi          = {10.1109/CVPR52734.2025.01235},
  bibsource    = {dblp computer science bibliography, https://dblp.org}
}

@inproceedings{fluxspace,
  author       = {Yusuf Dalva and
                  Kavana Venkatesh and
                  Pinar Yanardag},
  title        = {FluxSpace: Disentangled Semantic Editing in Rectified Flow Models},
  booktitle    = {{IEEE/CVF} Conference on Computer Vision and Pattern Recognition,
                  {CVPR} 2025, Nashville, TN, USA, June 11-15, 2025},
  pages        = {13083--13092},
  publisher    = {Computer Vision Foundation / {IEEE}},
  year         = {2025},
  url          = {https://openaccess.thecvf.com/content/CVPR2025/html/Dalva\_FluxSpace\_Disentangled\_Semantic\_Editing\_in\_Rectified\_Flow\_Models\_CVPR\_2025\_paper.html},
  doi          = {10.1109/CVPR52734.2025.01221},
  bibsource    = {dblp computer science bibliography, https://dblp.org}
}

@article{freeflux,
  author       = {Tianyi Wei and
                  Yifan Zhou and
                  Dongdong Chen and
                  Xingang Pan},
  title        = {FreeFlux: Understanding and Exploiting Layer-Specific Roles in RoPE-Based
                  MMDiT for Versatile Image Editing},
  journal      = {CoRR},
  volume       = {abs/2503.16153},
  year         = {2025},
  url          = {https://doi.org/10.48550/arXiv.2503.16153},
  doi          = {10.48550/ARXIV.2503.16153},
  eprinttype    = {arXiv},
  eprint       = {2503.16153},
  bibsource    = {dblp computer science bibliography, https://dblp.org}
}

@inproceedings{flashattn,
  author       = {Tri Dao and
                  Daniel Y. Fu and
                  Stefano Ermon and
                  Atri Rudra and
                  Christopher R{\'{e}}},
  editor       = {Sanmi Koyejo and
                  S. Mohamed and
                  A. Agarwal and
                  Danielle Belgrave and
                  K. Cho and
                  A. Oh},
  title        = {FlashAttention: Fast and Memory-Efficient Exact Attention with IO-Awareness},
  booktitle    = {Advances in Neural Information Processing Systems 35: Annual Conference
                  on Neural Information Processing Systems 2022, NeurIPS 2022, New Orleans,
                  LA, USA, November 28 - December 9, 2022},
  year         = {2022},
  url          = {http://papers.nips.cc/paper\_files/paper/2022/hash/67d57c32e20fd0a7a302cb81d36e40d5-Abstract-Conference.html},
  bibsource    = {dblp computer science bibliography, https://dblp.org}
}
}


\onecolumn

\renewcommand\thefigure{S\arabic{figure}}    
\setcounter{figure}{0}  
\renewcommand{\thesection}{\Alph{section}}
\setcounter{section}{0}
\renewcommand{\thetable}{S\arabic{table}}
\setcounter{table}{0}
\renewcommand{\theequation}{S\arabic{equation}}
\setcounter{equation}{0}
\renewcommand{\thealgorithm}{S\arabic{algorithm}}
\setcounter{algorithm}{0}

\clearpage
\appendix
\setcounter{page}{1}

\begin{center}
    {\LARGE \textbf{Group Relative Attention Guidance for Image Editing} \\[6pt]
    \textit{Supplementary Material}}
\end{center}
\vspace{1.0cm}

\begingroup
\setlength{\parskip}{6pt}
\hypersetup{linkcolor=black}
\startcontents[appendices]
\printcontents[appendices]{l}{1}{\section*{Contents}\setcounter{tocdepth}{2}}
\endgroup

\clearpage

\section{Pytorch Implementation of GRAG}
\label{sec:pytorch_impl}

\noindent
The proposed \textbf{Group Relative Attention Guidance (GRAG)} can be seamlessly integrated into existing DiT-based image editing models with only a few lines of code modification. 
Below, we provide an example implementation of GRAG based on a typical MM-Attention block from the \texttt{Diffusers} library in PyTorch.

\lstset{escapeinside={(*@}{@*)}}

\begin{lstlisting}[style=pyclean, caption={Implementation Code of GRAG}]
# Apply RoPE
if image_rotary_emb is not None:
    img_freqs, txt_freqs = image_rotary_emb
    img_query = apply_rotary_emb_qwen(img_query, img_freqs, use_real=False)
    img_key = apply_rotary_emb_qwen(img_key, img_freqs, use_real=False)
    txt_query = apply_rotary_emb_qwen(txt_query, txt_freqs, use_real=False)
    txt_key = apply_rotary_emb_qwen(txt_key, txt_freqs, use_real=False)

# Apply GRAG scaling
(*@\textcolor{red}{s\_idx, e\_idx, bias\_scale, delta\_scale = 4096, 8192, 1.0, 1.05}@*)
(*@\textcolor{red}{group\_bias = img\_key[:, s\_idx:e\_idx, :, :].mean(dim=1)}@*)
(*@\textcolor{red}{img\_key[:, s\_idx:e\_idx, :, :] = bias\_scale * group\_bias +}@*)
(*@\textcolor{red}{~~~~~~~~~~~~~~~~~~~~~~~~~~~~~~~delta\_scale * (img\_key[:, s\_idx:e\_idx, :, :] - group\_bias)}@*)

# Joint attention computation
joint_query = torch.cat([txt_query, img_query], dim=1)
joint_key = torch.cat([txt_key, img_key], dim=1)
joint_value = torch.cat([txt_value, img_value], dim=1)

joint_hidden_states = dispatch_attention_fn(
    joint_query,
    joint_key,
    joint_value,
    attn_mask=attention_mask,
    dropout_p=0.0,
    is_causal=False,
    backend=self._attention_backend,
)
\end{lstlisting}

\clearpage
\section{Theoretical Analysis of Group Relative Attention Guidance}

\subsection{Toy Experiment of the GRAG Theoretical Analysis}
\label{subsec:toy-validation}

To further validate the theoretical analysis presented in the main paper and Section~\ref{sec:GRAG}, we design a controlled toy experiment that isolates the effect of Group Relative Attention Guidance (GRAG) on attention score modulation. The aim is to empirically verify how adjusting the deviation scaling factor $\delta$ and the bias scaling factor $\lambda$ influences the distribution of attention across token groups.

\paragraph{Experimental Setup.}We simplify the MM-Attention mechanism in a minimal setting consisting of a single query $q = 1$ and three key tokens. 
\begin{equation}
    A(q,k^i)=\frac{e^{\langle q,k^i\rangle}}{e^{\langle q,k_s^{1}\rangle}+e^{\langle q,k_s^{2}\rangle}+e^{\langle q,k_t\rangle}},
\end{equation}
where $k_s^{1} = 1.9$ and $k_s^{2} = 1.2$ denote two source-image tokens and $k_t = 3.4$ denotes the editing-text token. Following the setup in Section \ref{sec:GRAG} of the main paper, two tokens form the \emph{source image token group}, while the remaining token represents the \emph{editing text token}. The raw inner-product responses $\langle q, k^i \rangle$ of these tokens serve as the unmodulated attention logits. This abstraction preserves the structural essence of text--image cross-attention while allowing us to precisely examine how GRAG affects token-wise attention allocation. Ideally, when increasing editing strength toward higher image–content consistency, the model should \emph{maintain responsiveness to the editing instruction} while \emph{smoothly increasing} the contribution of selected source-image tokens.

We compare four representative modulation strategies:
\begin{itemize}
    \setlength{\leftskip}{25pt} 
    \item Attention Weight - $\gamma$: Directly scaling attention scores after the Softmax layer.
    \item GRAG - $\lambda$: Modulating only the bias component while holding the deviation term fixed.
    \item GRAG - $\lambda, \delta$: Jointly scaling both the bias component and token-level deviations.
    \item GRAG - $\delta$: Modulating only the deviation component while keeping the bias fixed.
\end{itemize}

\begin{figure*}[h]
\centering
\includegraphics[width=\linewidth]{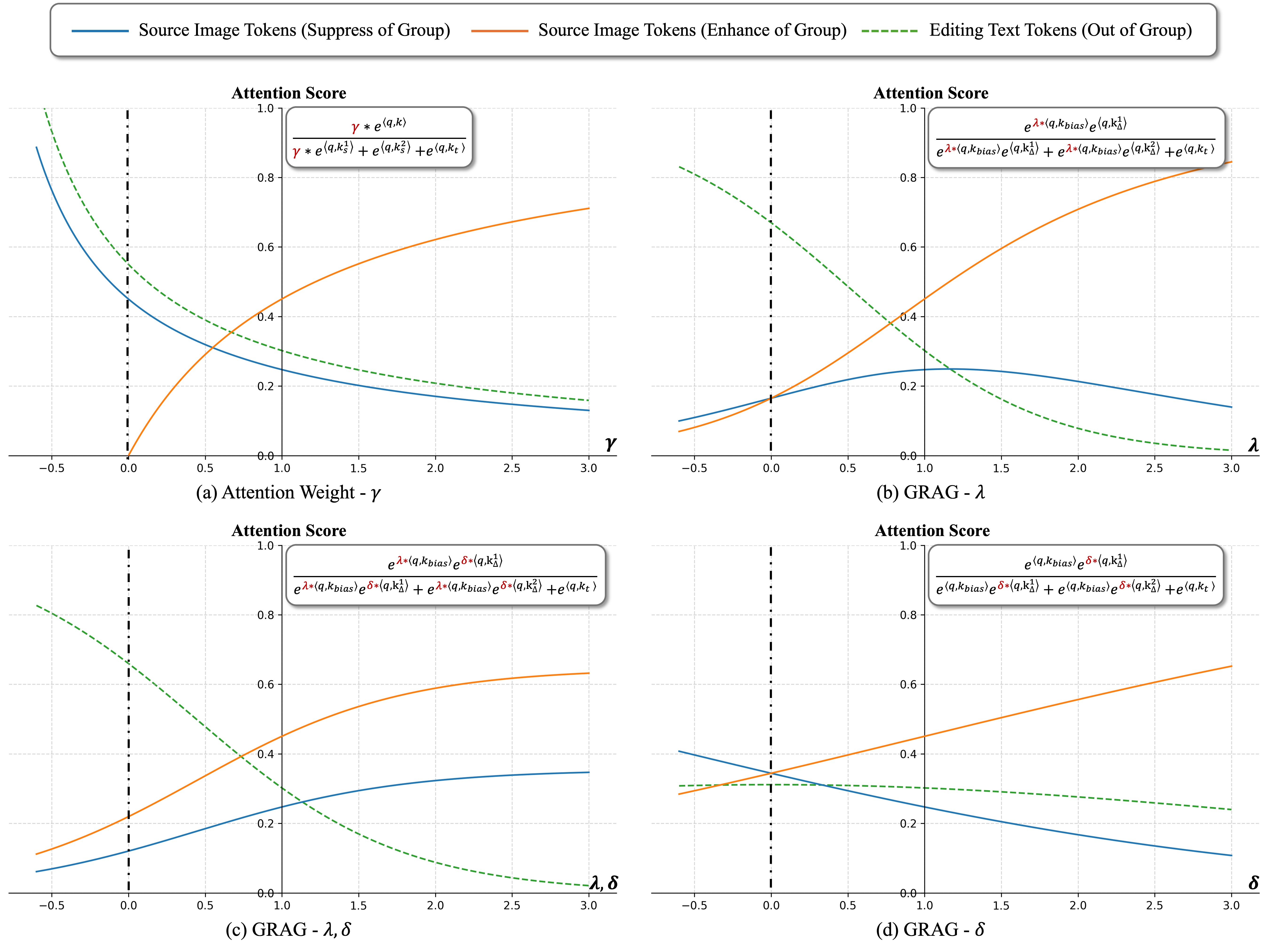}
\caption{
\textbf{Toy Experiment:} Effects of different attention modulation strategies on the attention scores within MM-Attention. GRAG-$\delta$ enhances or suppresses image tokens while preserving responsiveness to the editing instruction, enabling continuous and precise control over editing strength. The \textit{left vertical line} marks the lower modification boundary, and the \textit{$y$-axis} denotes the baseline without any modulation.}
\label{fig:supp-attn-analysis}
\end{figure*}

\paragraph{Experimental Results.}Figure~\ref{fig:supp-attn-analysis} illustrates this desired behavior using a toy example. The horizontal axis denotes the respective modulation parameter, and the vertical axis shows the post-Softmax attention scores of three token groups: suppressed source-image tokens, enhanced source-image tokens, and editing-text tokens. As seen in subfigures~(a)--(c), increasing $\gamma$ or $\lambda$, or jointly tuning $\lambda$ and $\delta$, rapidly suppresses attention to the editing-text tokens, causing the editing instruction to collapse and resulting in artifacts or editing failure despite increased attention to the image tokens.

In contrast, deviation-only modulation via GRAG-$\delta$ preserves stable attention on editing-text tokens while smoothly adjusting the relative emphasis among source-image tokens. This yields the most continuous and controllable editing-strength transition, free of undesirable discontinuities or instruction loss. Overall, GRAG-$\delta$ achieves the desired balance between instruction following and image consistency, providing the most reliable and precise editing-strength control among all tested strategies.

\clearpage
\subsection{Relationship Between Attention Entropy and GRAG}
\label{subsec:attn-entropy}

To quantify how GRAG influences the distribution of attention inside MM-Attention, we measure the \emph{attention entropy} of each attention map. Given an attention distribution $A^{(l)} \in \mathbb{R}^{N}$ from layer $l$, its entropy is defined as:
\begin{equation}
H(A^{(l)}) = - \sum_{i=1}^{N} A^{(l)}_i \log A^{(l)}_i,
\label{eq:entropy}
\end{equation}
where lower entropy indicates a more concentrated attention pattern and stronger focus on specific tokens.

\begin{figure}[h]
\centering
\includegraphics[page=1,width=\linewidth]{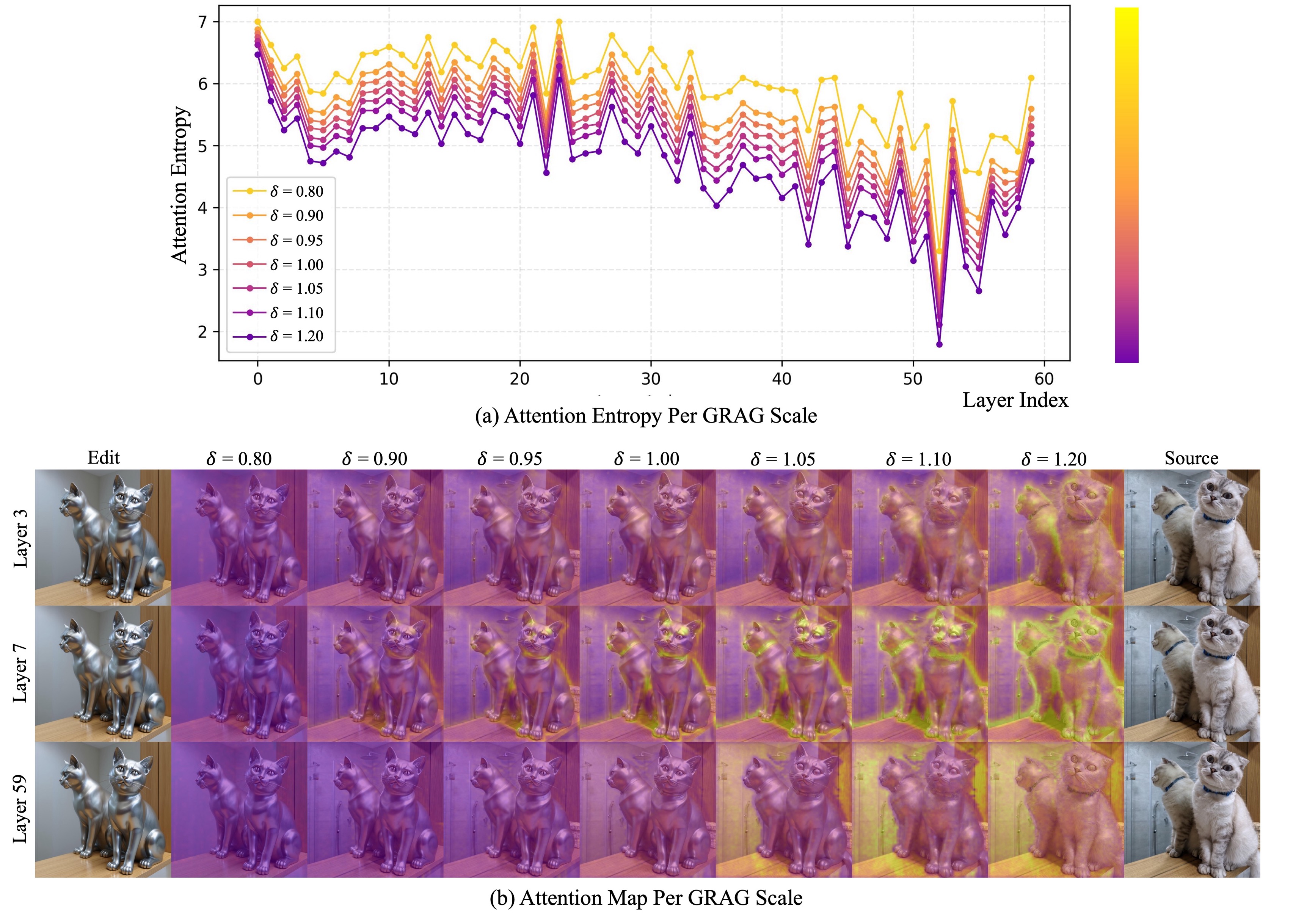}
\caption{
Relationship between attention entropy and GRAG. (a) Increasing the GRAG value on source image tokens leads to a monotonic decrease in attention entropy across layers. (b) The query’s attention over source image tokens becomes progressively more concentrated as the GRAG value increases.
}

\label{fig:supp-attn-entropy}
\vspace{-7mm}
\end{figure}

We compute attention entropy for all layers of the Qwen-Image-Edit model under different GRAG values. Specifically, for each GRAG scale, we extract the attention maps associated with the source-image tokens and compute their entropies using Equation~\eqref{eq:entropy}. The aggregated results are shown in Figure~\ref{fig:supp-attn-entropy}(a). As the GRAG value increases, the attention entropy in every layer decreases monotonically, indicating that GRAG strengthens the model’s focus on conditioning information and thereby increases the effective editing strength.

To provide a more intuitive interpretation, we visualize the attention maps corresponding to the same settings. As shown in Figure~\ref{fig:supp-attn-entropy}(b), increasing the GRAG value causes the query’s attention to progressively concentrate on the original cat contours. This focused attention corresponds to stronger preservation of source-image structure, leading to improved consistency between the edited output and the reference image.

These results collectively demonstrate that GRAG modulates editing strength by adjusting the concentration of cross-attention, aligning the empirical observations with the theoretical analysis presented in the main paper.

\clearpage
\subsection{GRAG of Source Image Tokens}
\label{subsec:image-bias-delta}

\begin{figure}[h]
\vspace{-3mm}
\centering
\includegraphics[page=1,width=\linewidth]{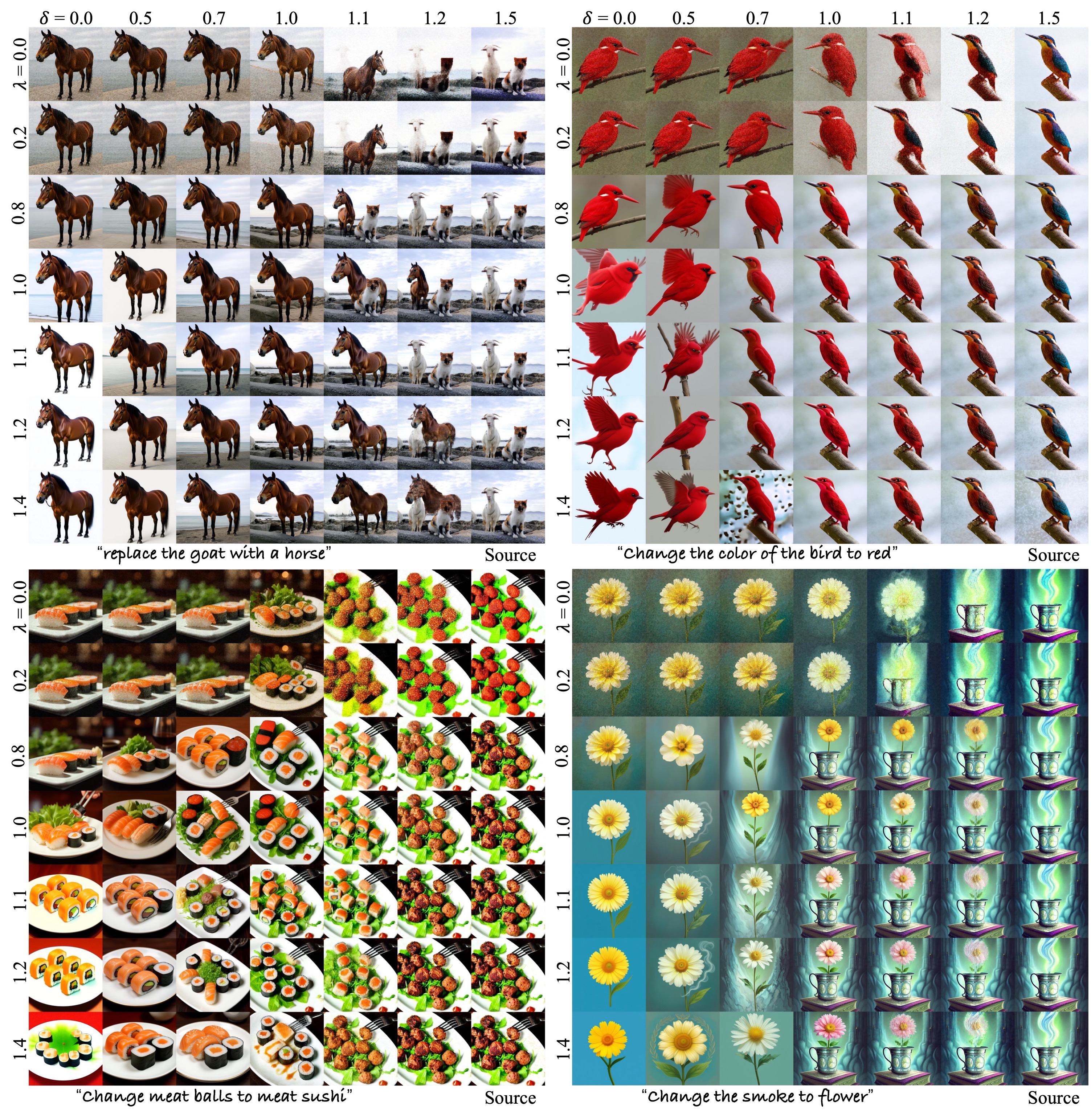}
\vspace{-8mm}
\caption{Effects of the bias and variation components in image-key embeddings.} 
\label{fig:supp-ablation-img}
\vspace{-7mm}
\end{figure}

To further understand the distinct roles of the bias and variation components in GRAG, we separately perform continuous adjustments on the text-embedding scaling factors $\lambda$ (bias scaling) and $\delta$ (variation scaling). 
\begin{equation}
\hat{k}_\mathrm{s}^i = \lambda \cdot k_{\text{bias}} + \delta \cdot \left(k_\mathrm{s}^i - k_{\text{bias}}\right)
\end{equation}
The results are shown in Fig.~\ref{fig:supp-ablation-img}. When only the bias component is modulated, the overall editing effect remains relatively stable, yet noticeable changes in image quality occur. In contrast, when only the variation component is adjusted, the semantic content of the edited image changes rapidly with increasing editing strength, while the overall image quality remains largely consistent. These observations support our interpretation that the bias component corresponds to the model’s intrinsic editing behavior, whereas the variation component encodes the actual content-specific editing signals.

\clearpage
\subsection{GRAG of Editing Text Tokens}
\label{subsec:text-bias-delta}
\begin{figure}[h]
\centering
\includegraphics[page=1,width=\linewidth]{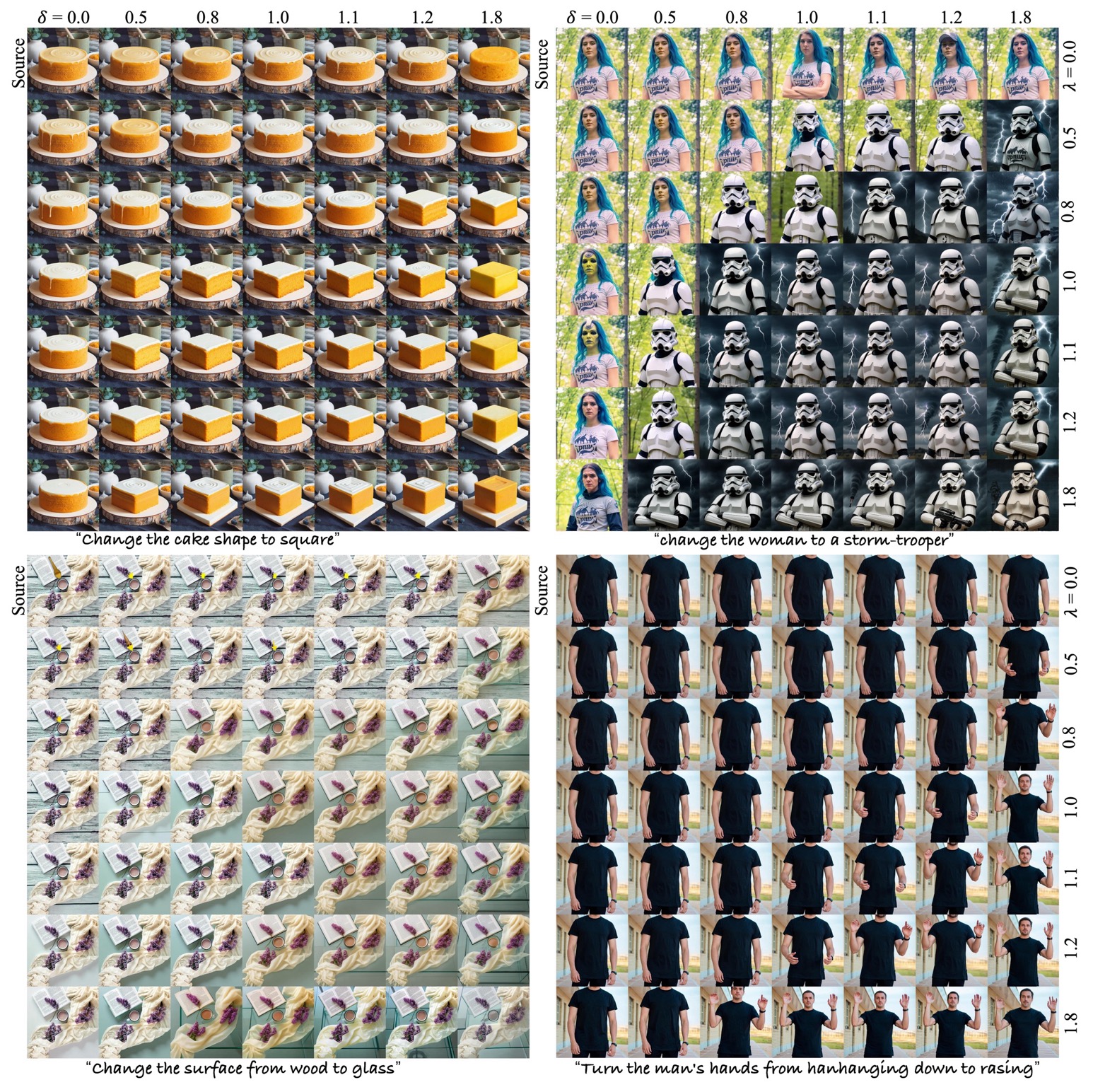}
\caption{Effects of the bias and variation components in text-key embeddings.} 
\label{fig:supp-ablation-txt}
\vspace{-7mm}
\end{figure}

We conduct the same decomposition-based analysis on the text tokens to examine the roles of the bias and variation components, with results shown in Fig.~\ref{fig:supp-ablation-txt}. Similar to the observations for image tokens, the bias component governs the model's inherent editing behavior, while the variation component drives the content-specific semantic changes. However, applying GRAG to text tokens reveals a noticeably stronger degree of semantic-level editing control, indicating that modulation of text embeddings provides a more direct and expressive pathway for manipulating editing intensity.

\clearpage
\section{Limitation \& Discussion}
\label{sec:limitation}
\begin{figure*}[h]
\vspace{-1mm}
\centering
\includegraphics[width=0.55\linewidth]{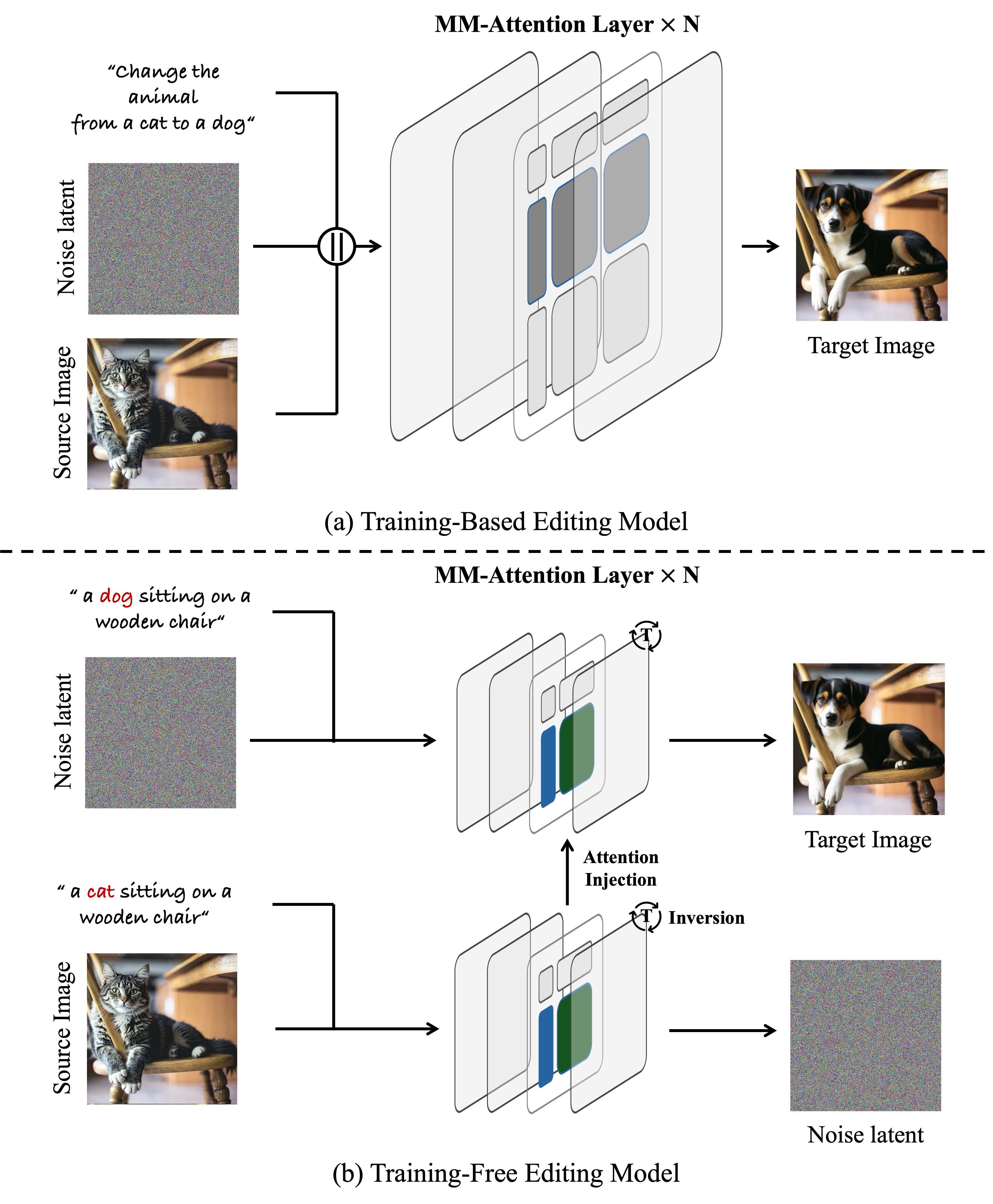}
\caption{Structural differences between training-based and training-free editing methods.} 
\label{fig:supp-limitation}
\end{figure*}

\begin{figure*}[h]
\vspace{-3mm}
\centering
\includegraphics[width=0.75\linewidth]{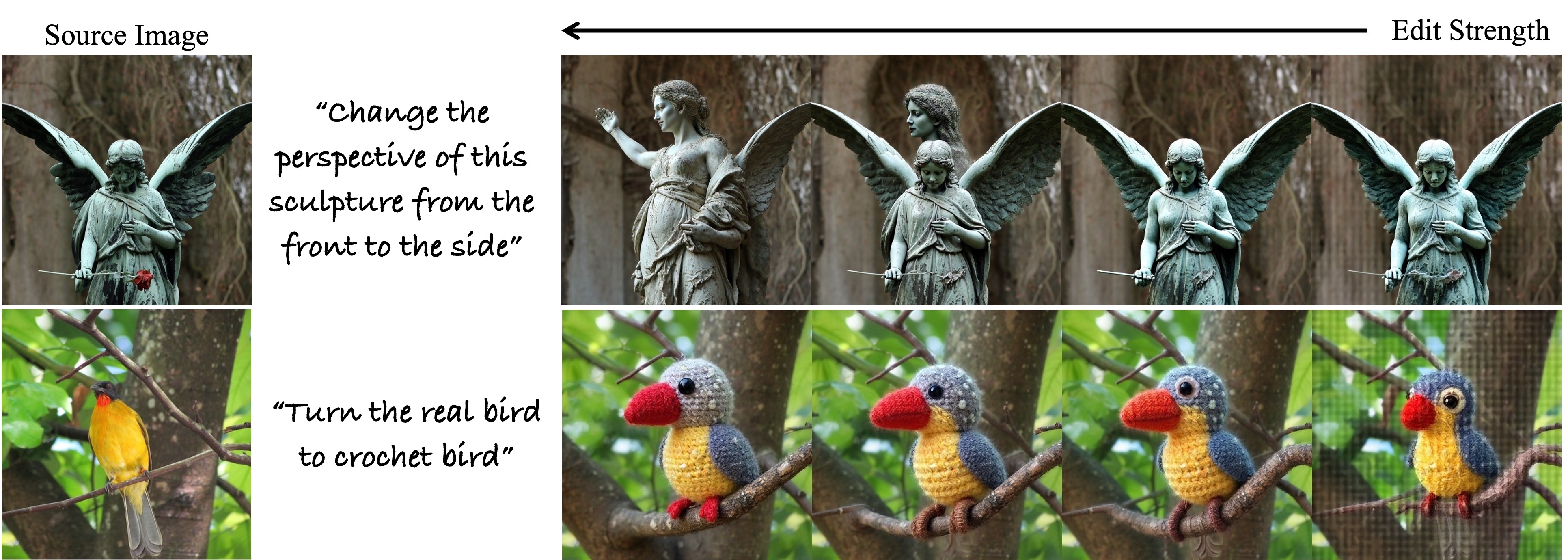}
\caption{Failure cases of applying GRAG to training-free editing methods.} 
\label{fig:supp-limitation-vis}
\end{figure*}

We observe that the effectiveness of GRAG is inherently limited when applied to training-free image editing methods. As illustrated in Fig.~\ref{fig:supp-limitation}, training-based TI2I models possess a unified architecture in which the editing instruction and source-image information interact within the same MM-Attention layers, yielding better compatibility with GRAG. In contrast, training-free approaches are built upon T2I models that are not originally designed for editing; they rely on additional inversion procedures and attention injection to approximate the interaction between edited and source-image features. This structural mismatch constrains the impact of GRAG and lead to degraded image quality or ineffective edits (Fig.~\ref{fig:supp-limitation-vis}).

\clearpage
\section{Additional Qualitative Results}
\label{sec:more-results}

\begin{figure*}[h]
\centering
\includegraphics[width=0.95\linewidth]{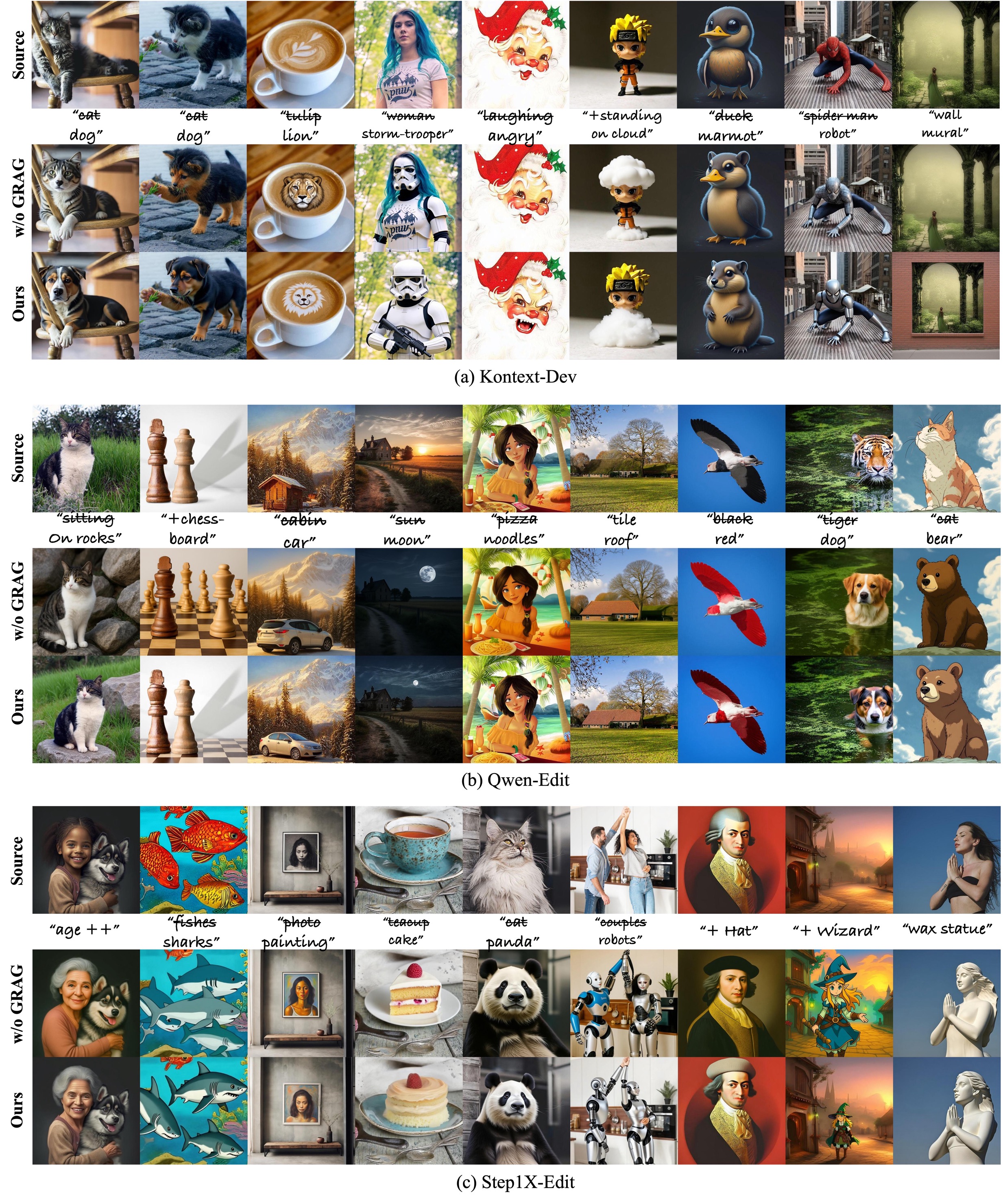}
\caption{Additional qualitative results on existing image editing models.} 
\label{fig:supp-comparsion-2}
\vspace{-5mm}
\end{figure*}

\begin{figure*}[h]
\vspace{-10mm}
\centering
\includegraphics[width=0.9\linewidth]{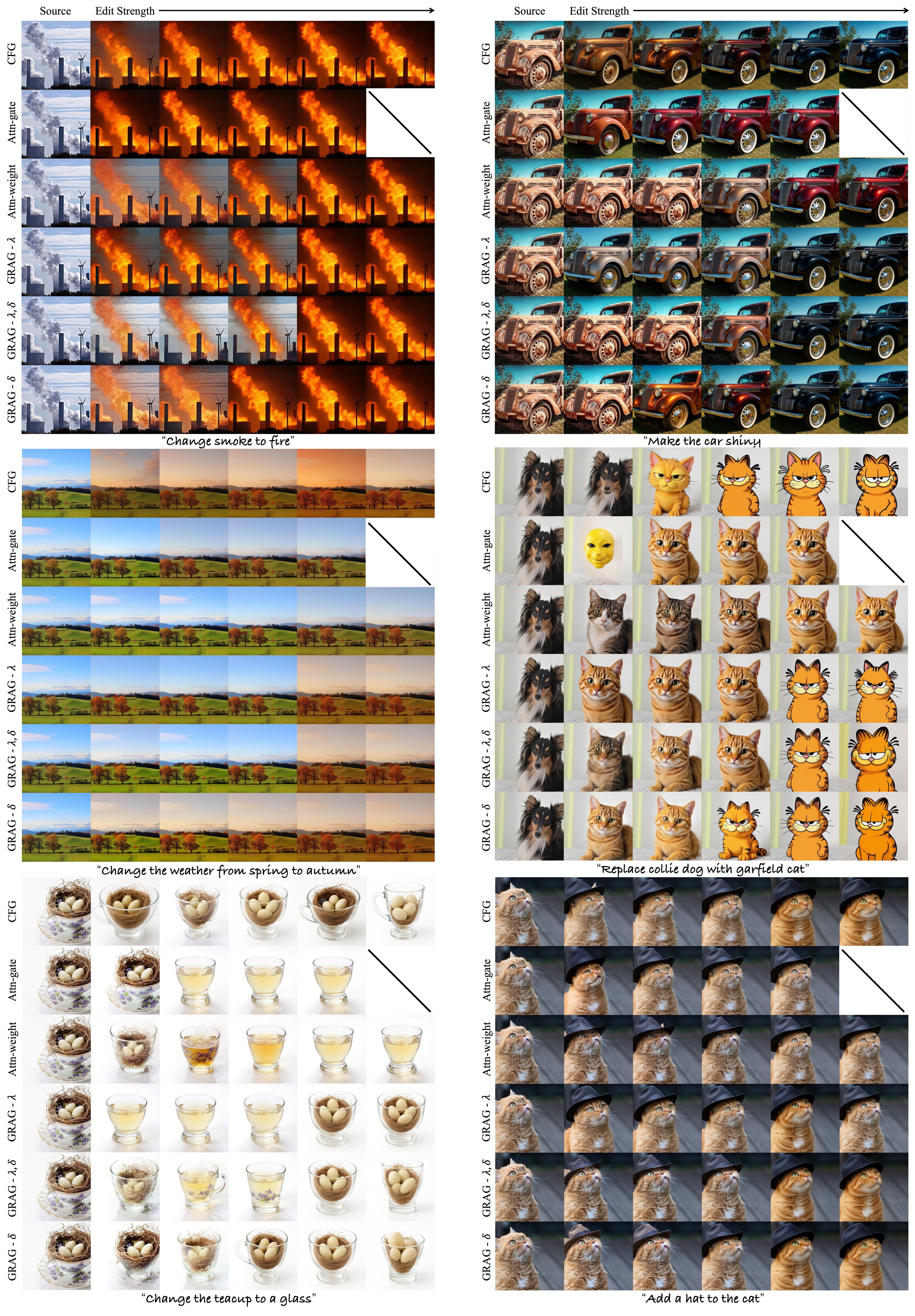}
\caption{Additional comparison results with existing editing-strength control methods.} 
\label{fig:supp-comparsion-1}
\vspace{-5mm}
\end{figure*}

\clearpage
\section{Additional Feature Visualization}
\label{sec:vis_more}

\begin{figure*}[h]
\centering
\includegraphics[width=0.95\linewidth]{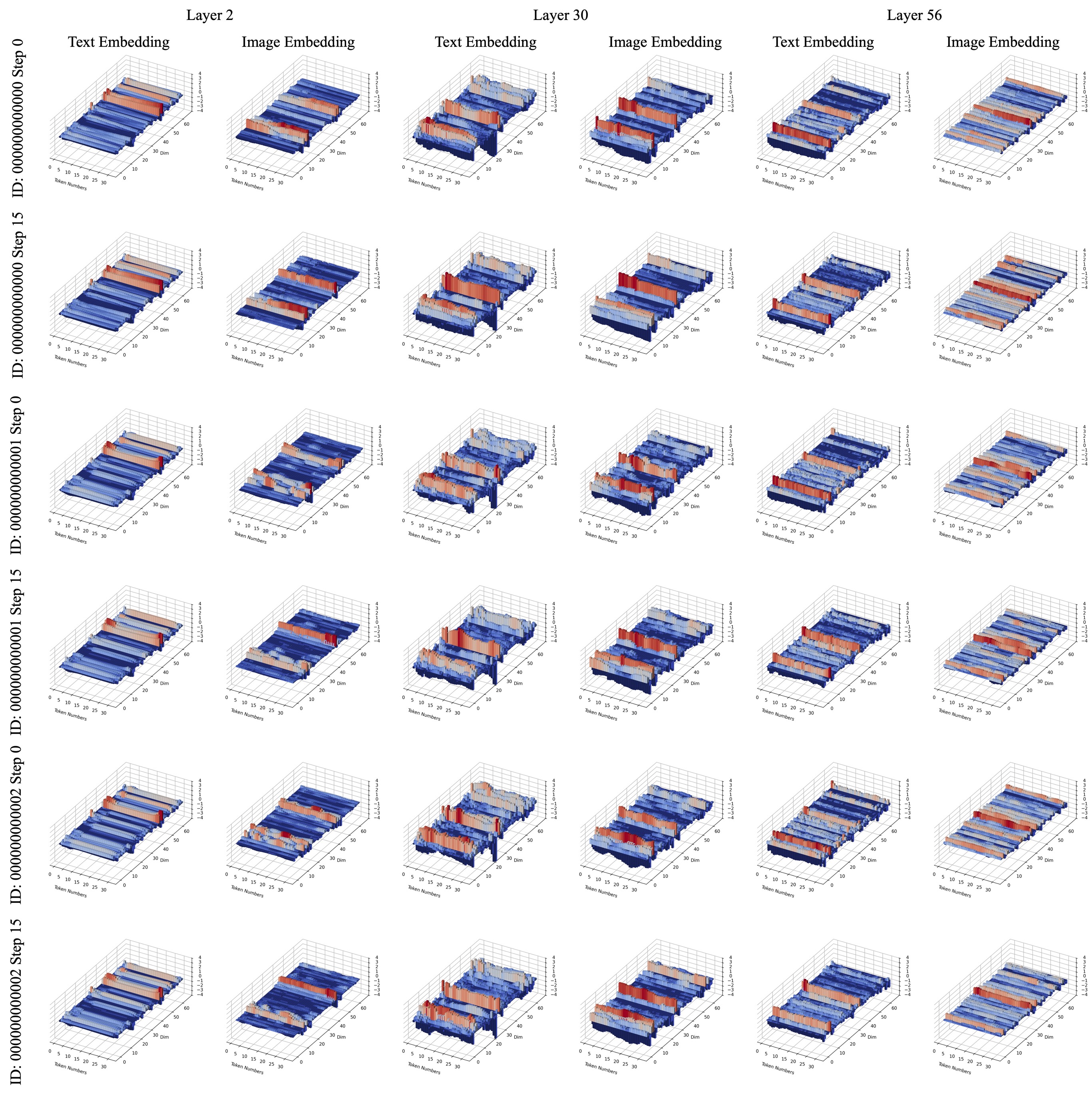}
\caption{Additional visualizations of text and image embedding features. 
Features within the same layer share similar distributions, indicating limited correlation with model inputs or denoising steps. 
Please zoom in to view finer details.} 
\label{fig:supp-vis-3d}
\vspace{-5mm}
\end{figure*}

\begin{figure*}[h]
\centering
\includegraphics[width=1.0\linewidth]{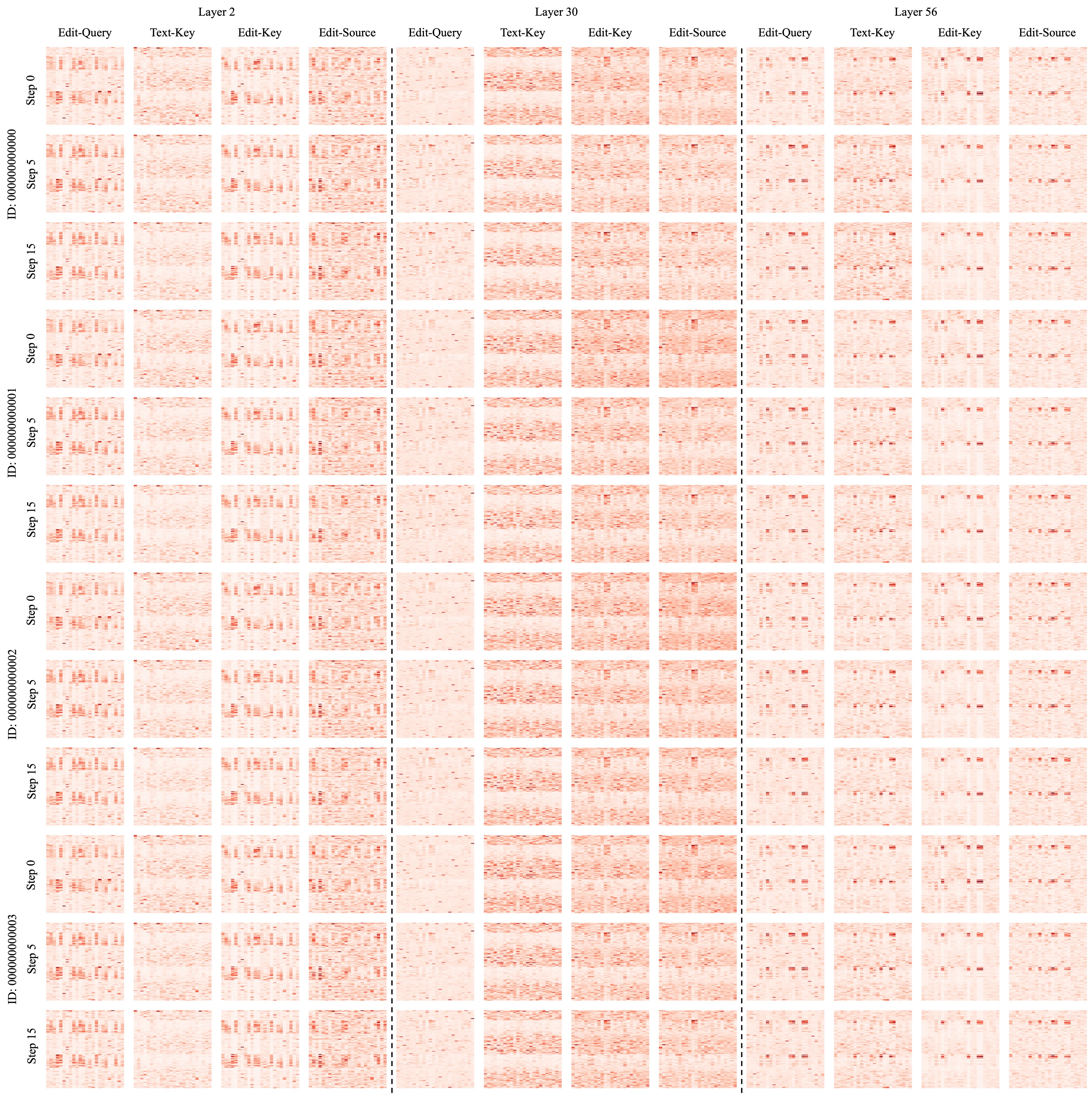}
\caption{Additional visualizations of aggregating different tokens along the sequence dimension.
Please zoom in to view finer details.} 
\label{fig:supp-vis-2d-1}
\vspace{-5mm}
\end{figure*}

\begin{figure*}[h]
\centering
\includegraphics[width=0.8\linewidth]{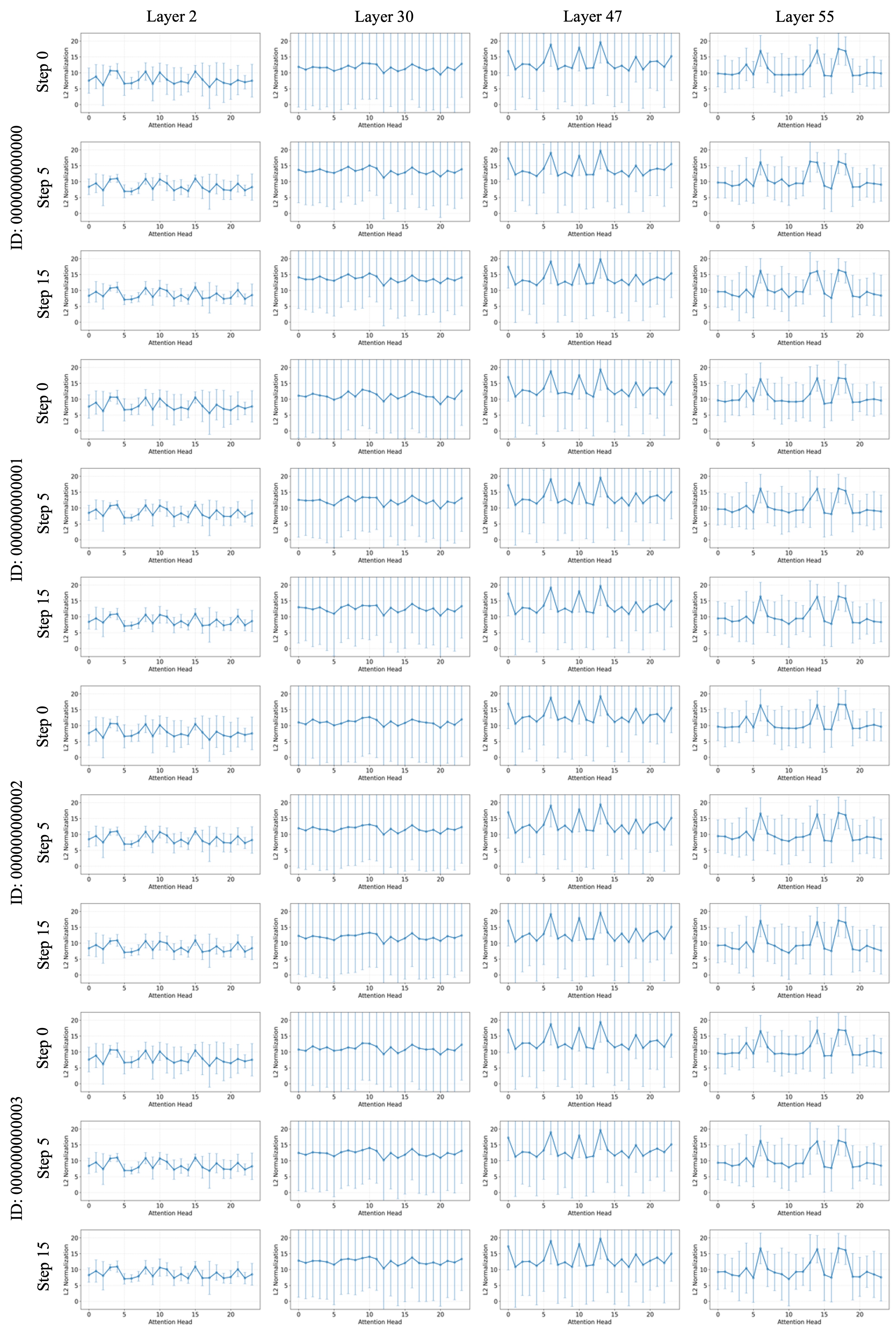}
\caption{Additional visualizations of Query-edit embedding mean vector magnitudes and standard deviations across different attention heads.
Please zoom in to view finer details.} 
\label{fig:supp-vis-2d-q-edit}
\vspace{-5mm}
\end{figure*}

\begin{figure*}[h]
\centering
\includegraphics[width=0.8\linewidth]{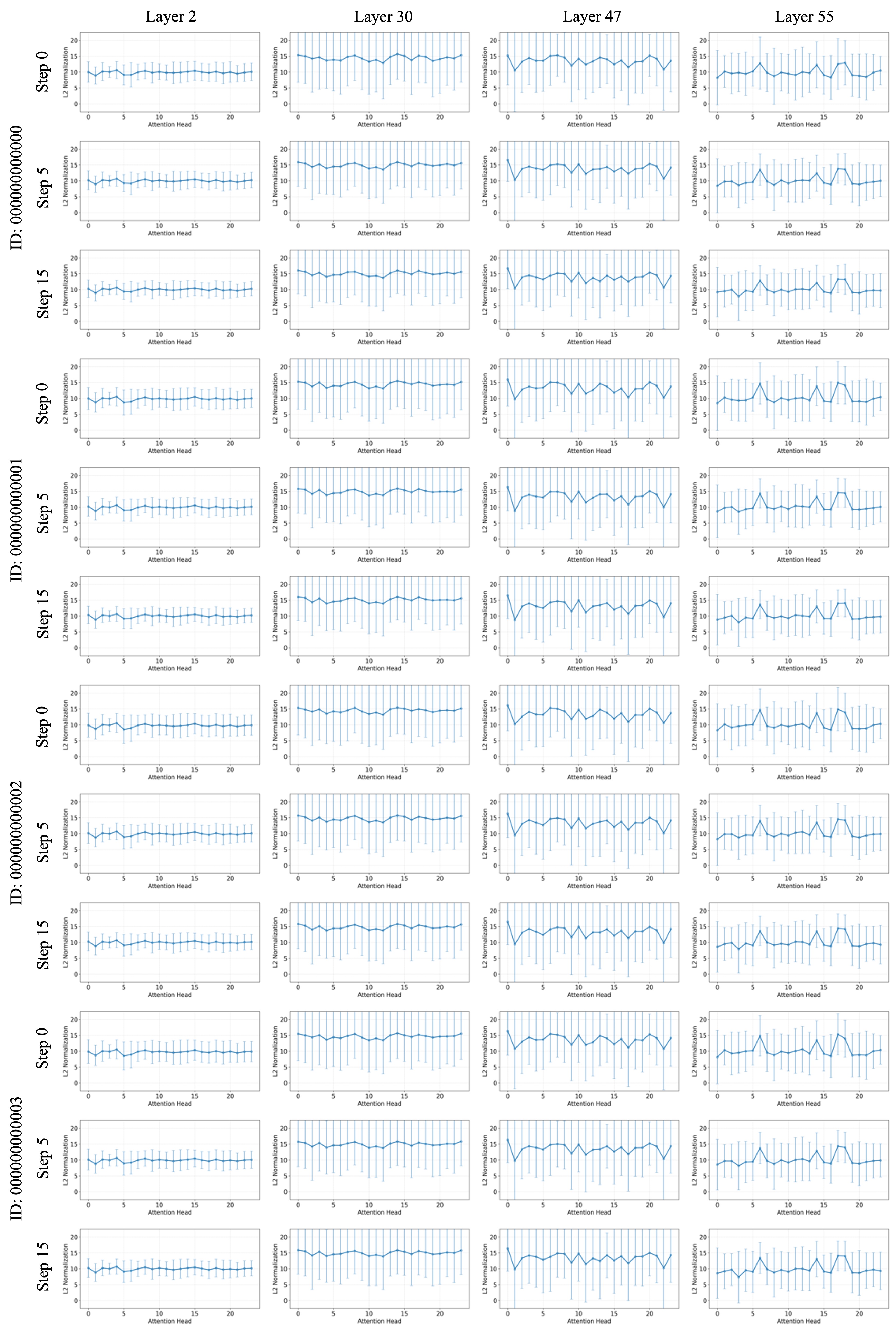}
\caption{Additional visualizations of Key-text embedding mean vector magnitudes and standard deviations across different attention heads.
Please zoom in to view finer details.} 
\label{fig:supp-vis-2d-k-txt}
\vspace{-5mm}
\end{figure*}

\begin{figure*}[h]
\centering
\includegraphics[width=0.8\linewidth]{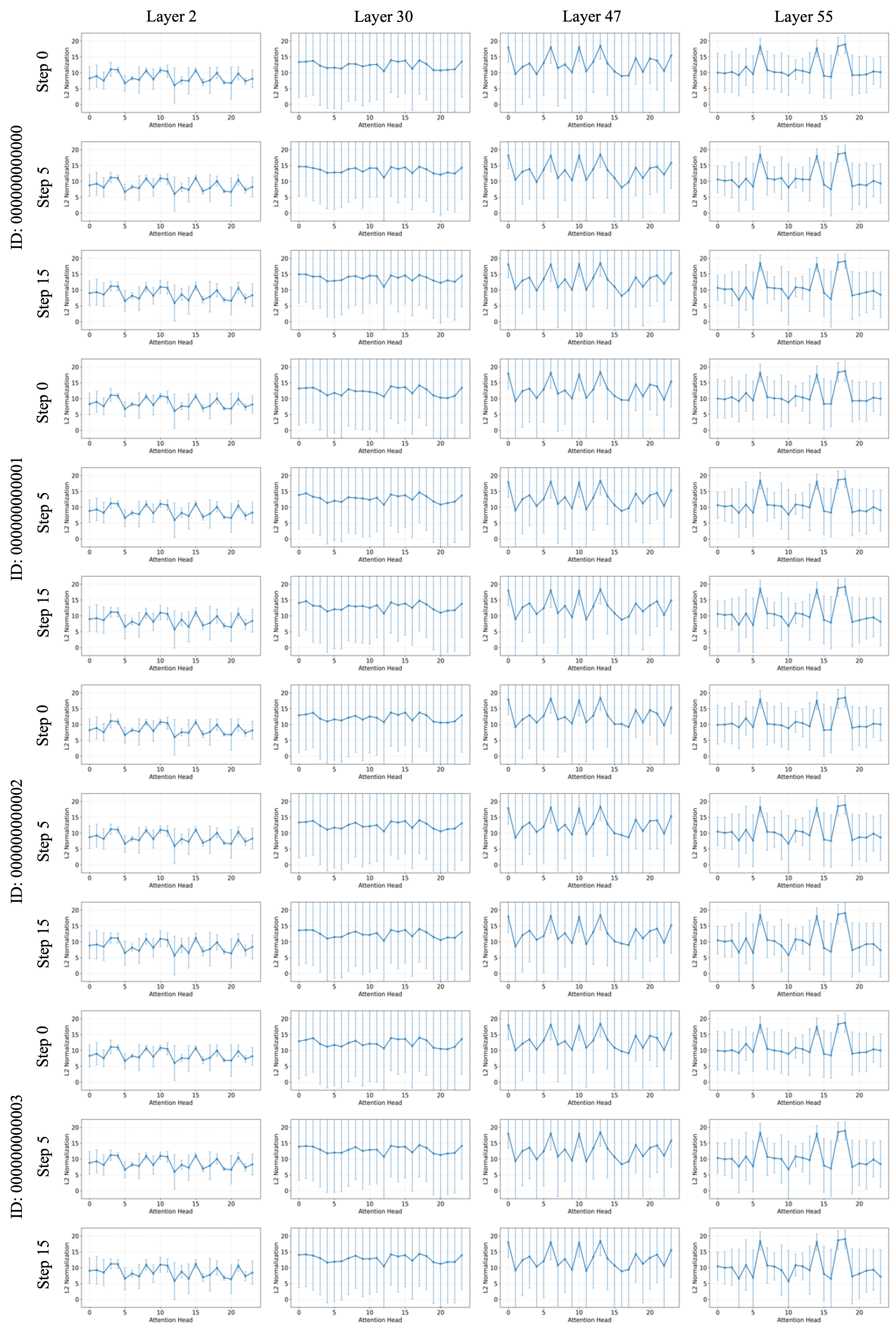}
\caption{Additional visualizations of Key-edit embedding mean vector magnitudes and standard deviations across different attention heads.
Please zoom in to view finer details.} 
\label{fig:supp-vis-2d-k-edit}
\vspace{-5mm}
\end{figure*}

\begin{figure*}[h]
\centering
\includegraphics[width=0.8\linewidth]{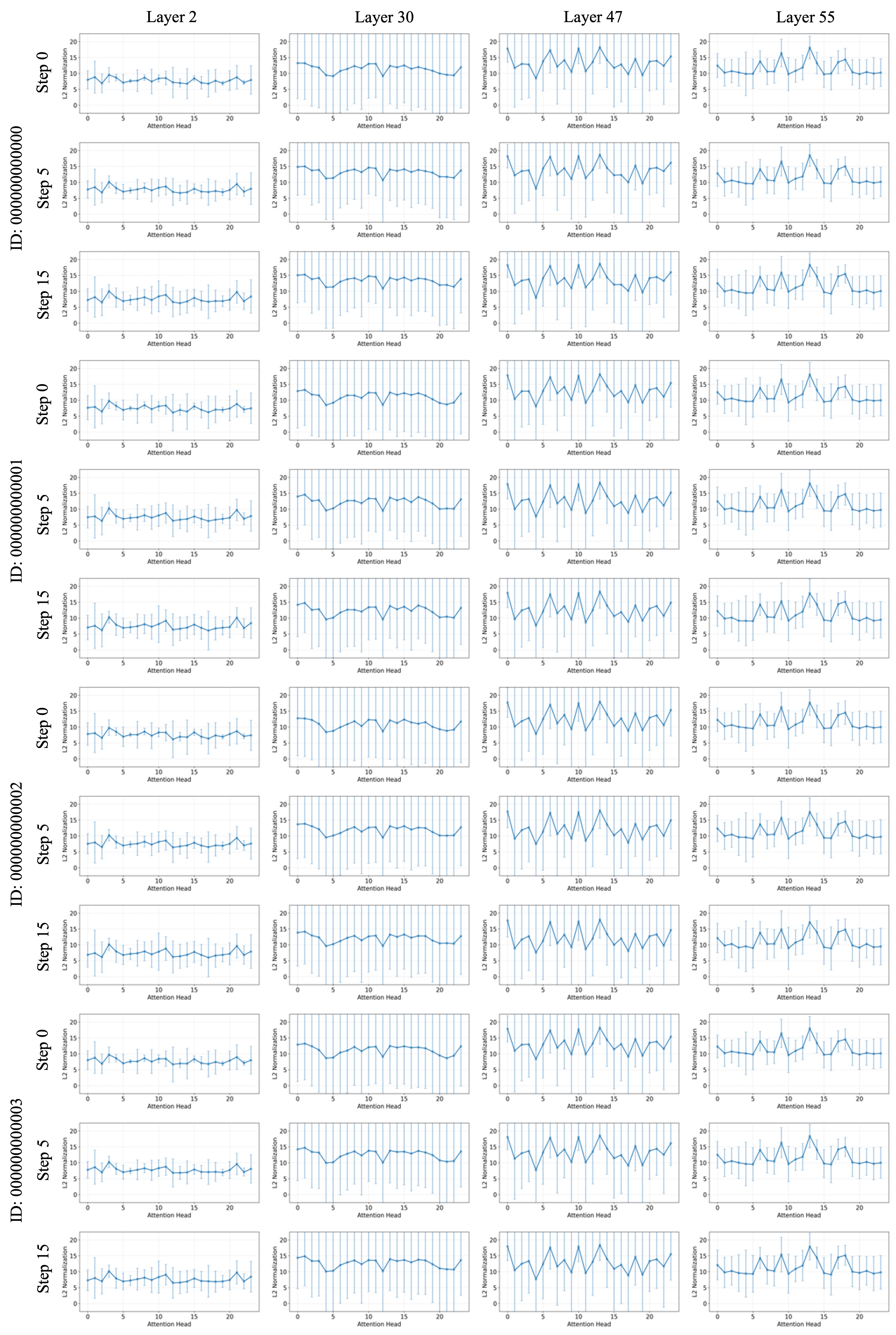}
\caption{Additional visualizations of Key-src embedding mean vector magnitudes and standard deviations across different attention heads.
Please zoom in to view finer details.} 
\label{fig:supp-vis-2d-k-src}
\vspace{-5mm}
\end{figure*}

\begin{figure*}[h]
\centering
\includegraphics[width=0.95\linewidth]{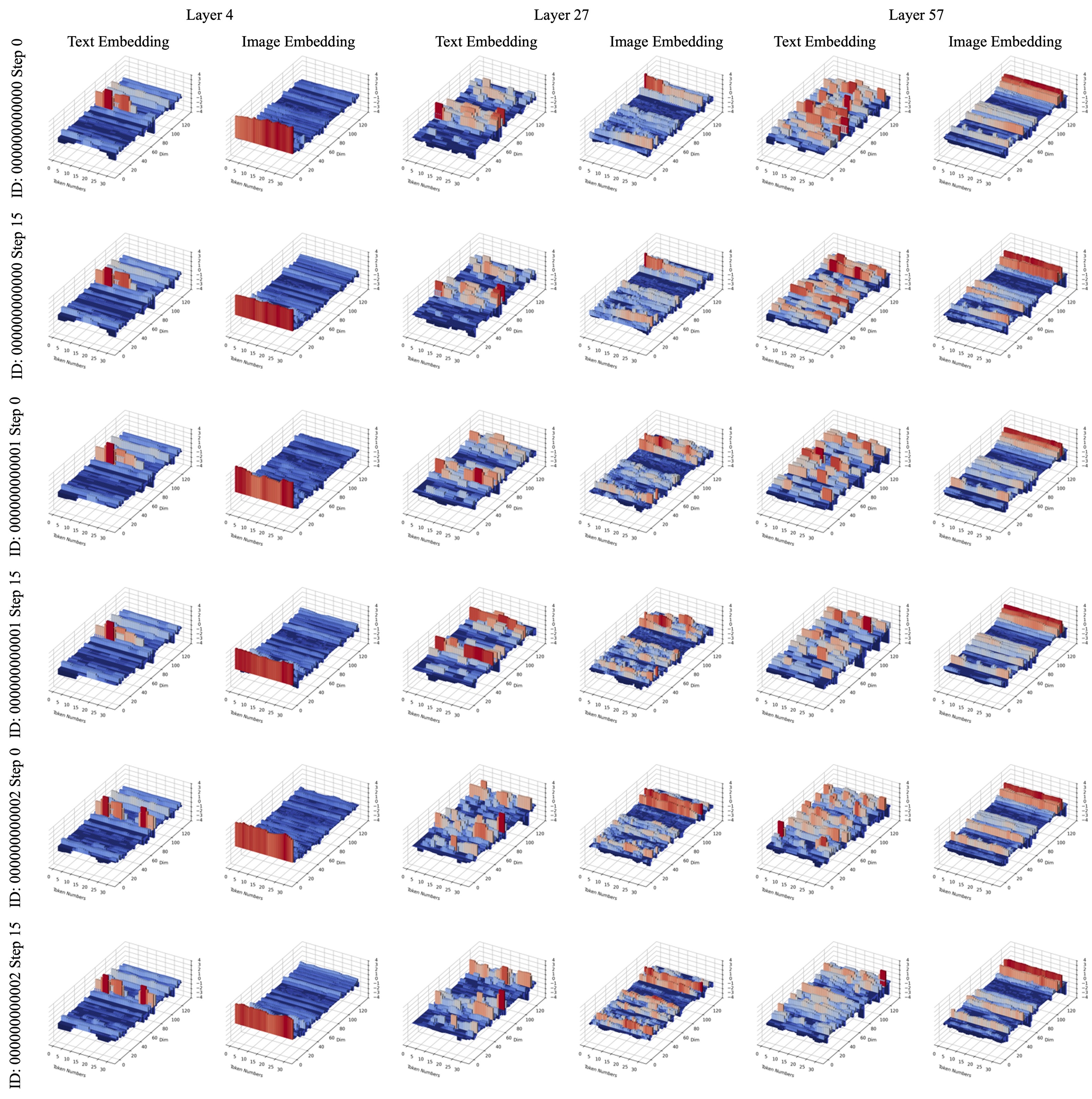}
\caption{Additional visualizations of text and image embedding features. 
Features within the same layer share similar distributions, indicating limited correlation with model inputs or denoising steps. 
Please zoom in to view finer details.} 
\label{fig:supp-vis-3d-qwen}
\vspace{-5mm}
\end{figure*}

\begin{figure*}[h]
\centering
\includegraphics[width=1.0\linewidth]{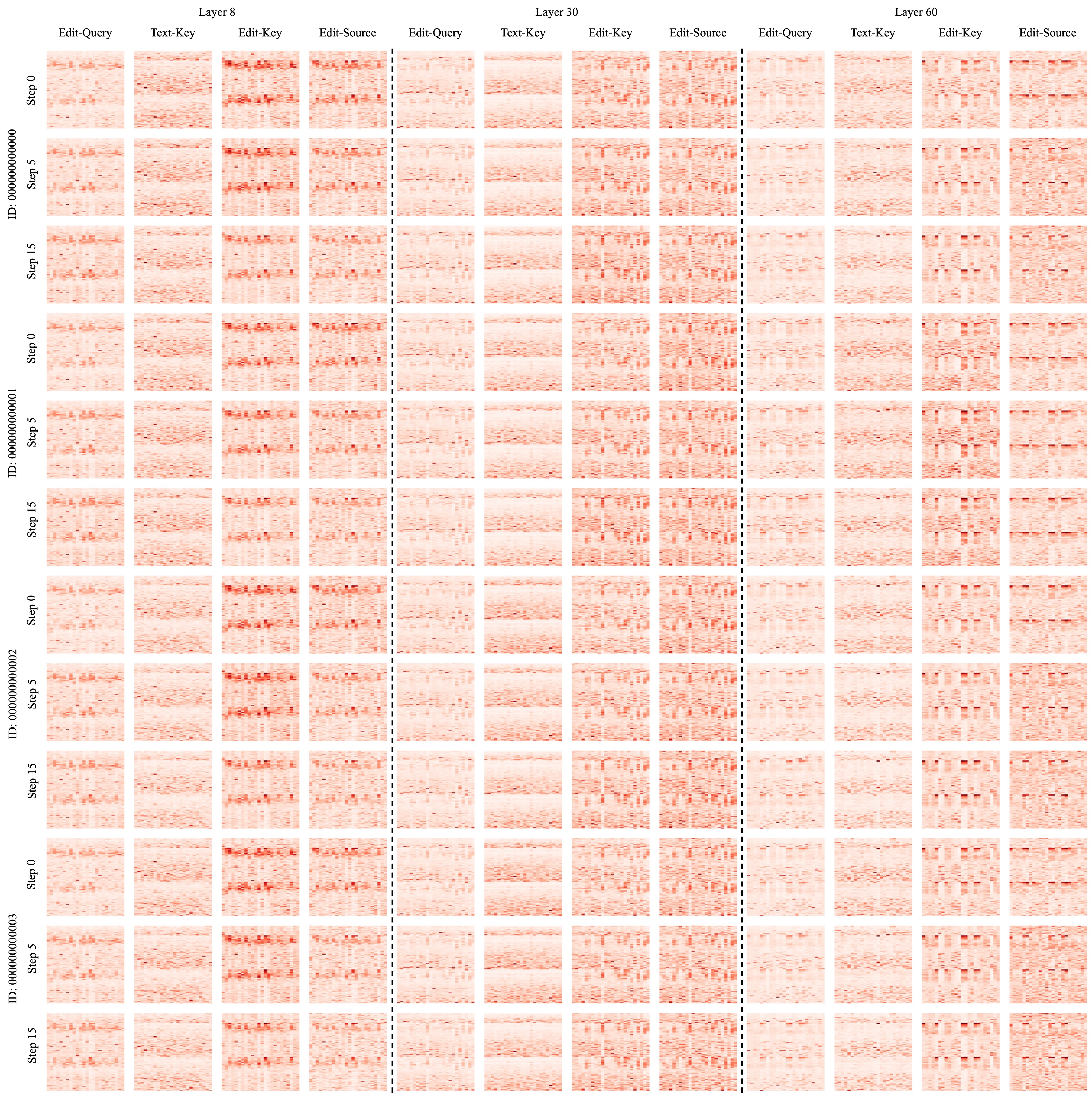}
\caption{Additional visualizations of aggregating different tokens along the sequence dimension.
Please zoom in to view finer details.} 
\label{fig:supp-vis-2d-1-qwen}
\vspace{-5mm}
\end{figure*}

\begin{figure*}[h]
\centering
\includegraphics[width=0.8\linewidth]{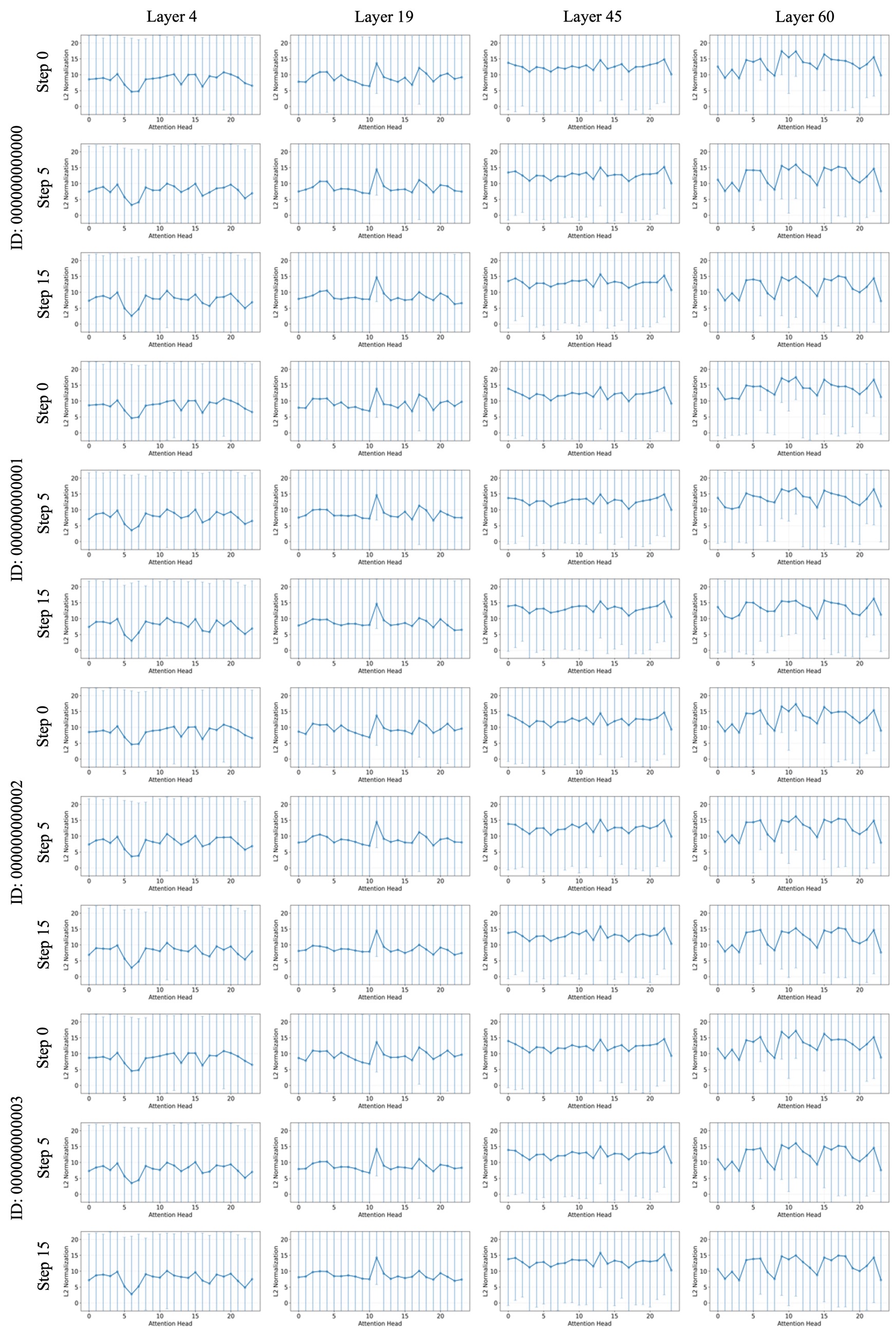}
\caption{Additional visualizations of Query-edit embedding mean vector magnitudes and standard deviations across different attention heads.
Please zoom in to view finer details.} 
\label{fig:supp-vis-2d-q-edit-qwen}
\vspace{-5mm}
\end{figure*}

\begin{figure*}[h]
\centering
\includegraphics[width=0.8\linewidth]{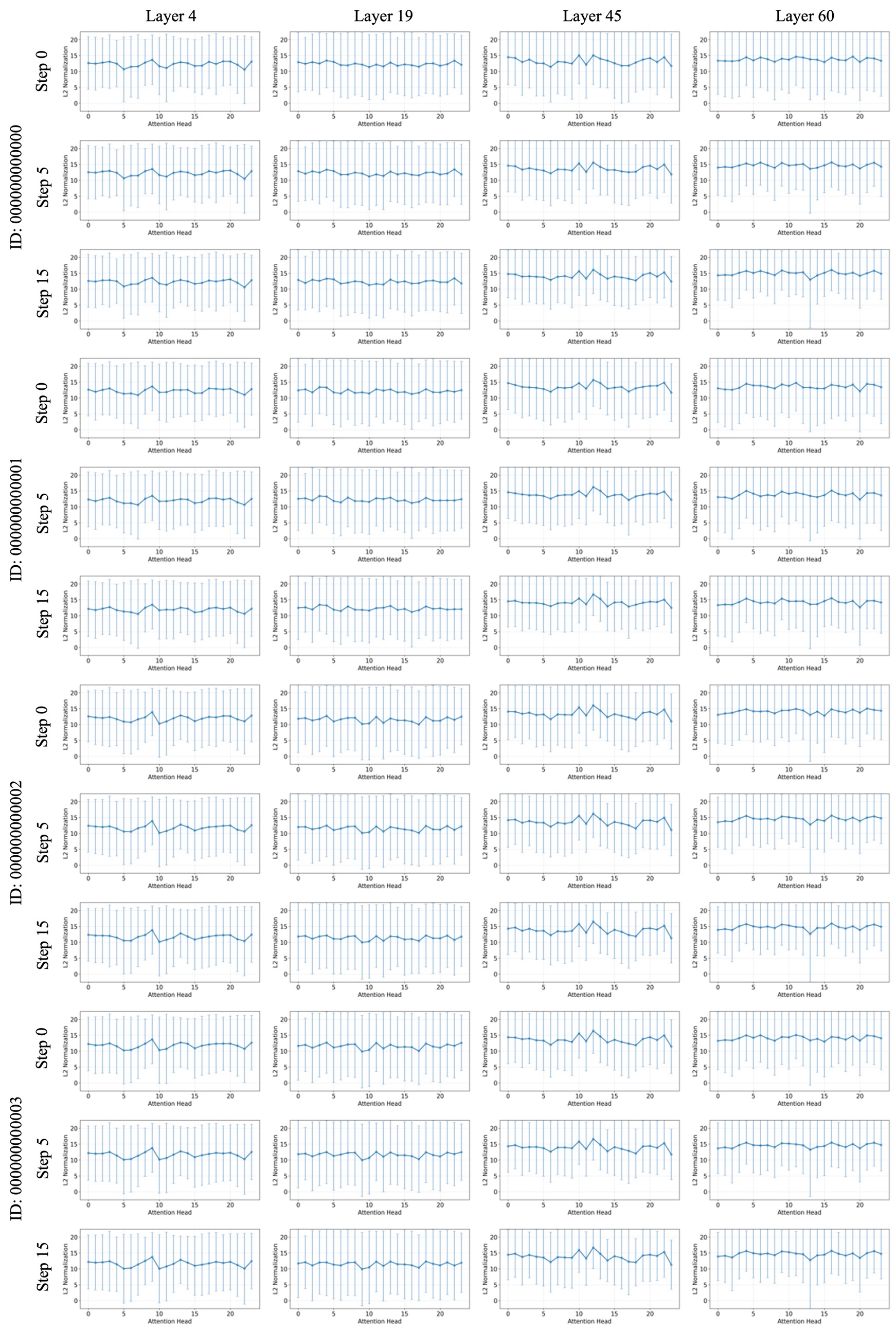}
\caption{Additional visualizations of Key-text embedding mean vector magnitudes and standard deviations across different attention heads.
Please zoom in to view finer details.} 
\label{fig:supp-vis-2d-k-txt-qwen}
\vspace{-5mm}
\end{figure*}

\begin{figure*}[h]
\centering
\includegraphics[width=0.8\linewidth]{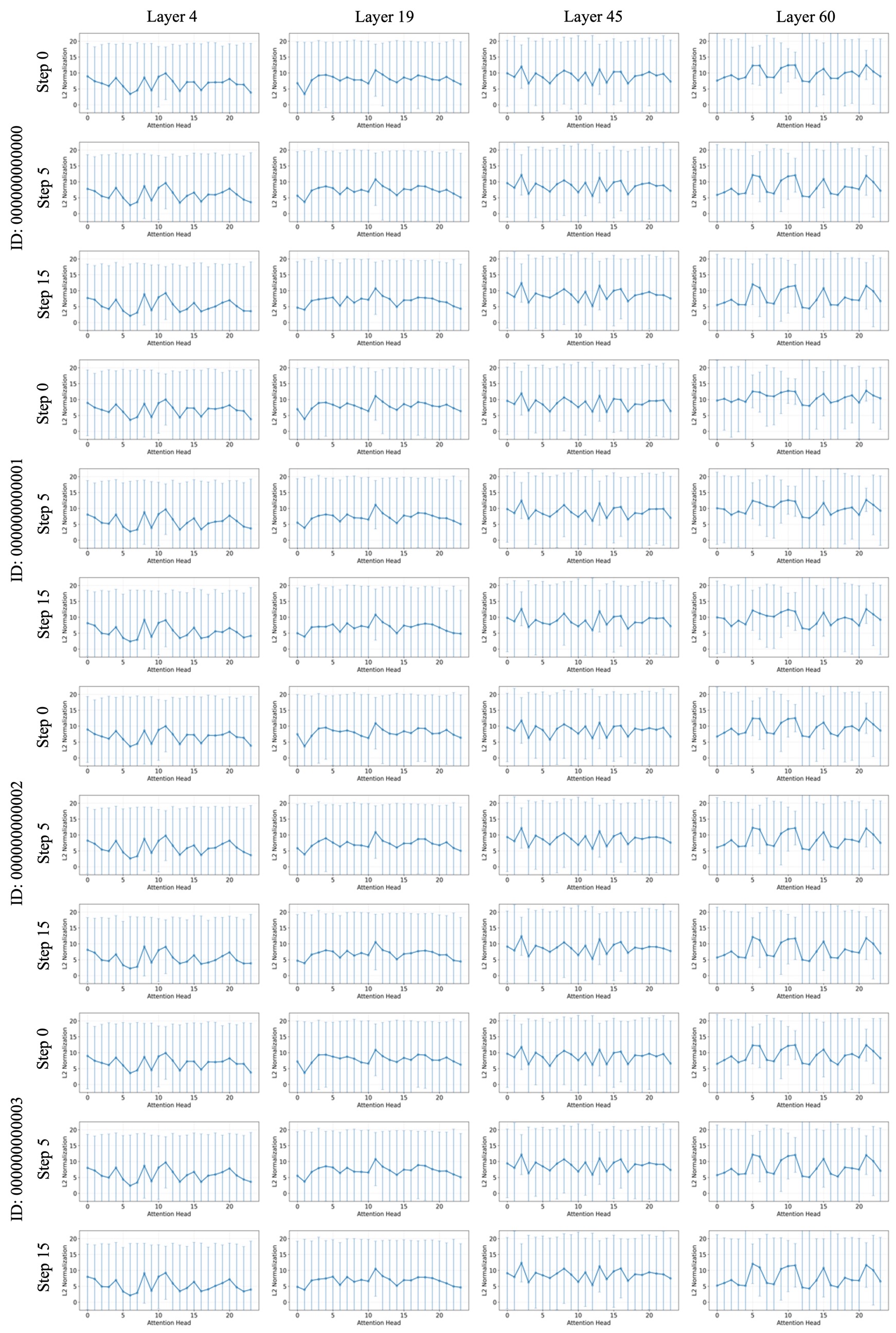}
\caption{Additional visualizations of Key-edit embedding mean vector magnitudes and standard deviations across different attention heads.
Please zoom in to view finer details.} 
\label{fig:supp-vis-2d-k-edit-qwen}
\vspace{-5mm}
\end{figure*}

\begin{figure*}[h]
\centering
\includegraphics[width=0.8\linewidth]{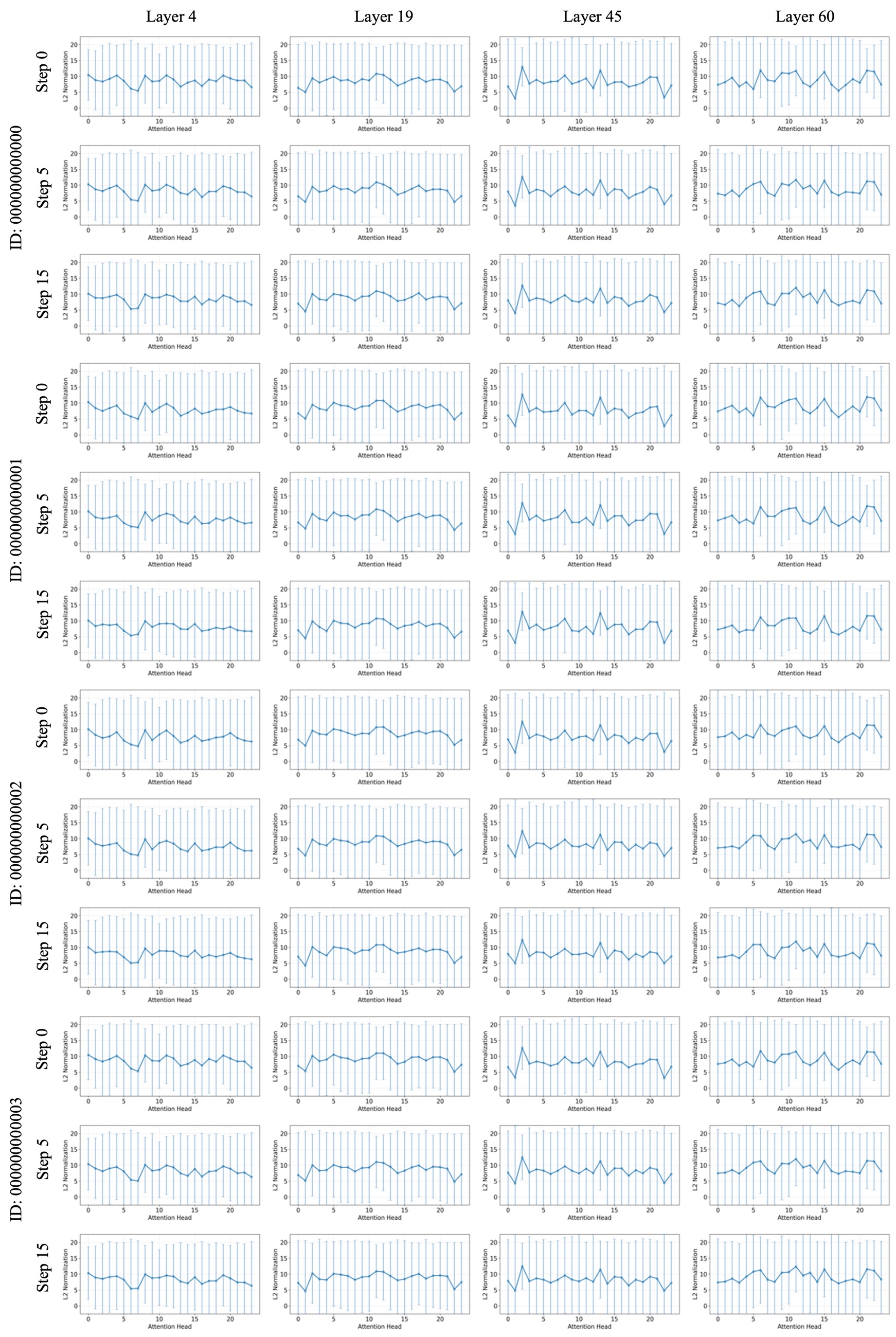}
\caption{Additional visualizations of Key-src embedding mean vector magnitudes and standard deviations across different attention heads.
Please zoom in to view finer details.} 
\label{fig:supp-vis-2d-k-src-qwen}
\vspace{-5mm}
\end{figure*}

\noindent
We provide additional Kontext and Qwen-Edit model feature distribution statistics corresponding to Figures~\ref{fig:intro-1}, \ref{fig:rethinking-1}, and \ref{fig:rethinking-2} in the main paper. 
Consistent with the experiments presented in the main text, we analyze different image editing samples (IDs) across various denoising steps and model layers to examine the correlation between feature distributions and these three factors. 
Figure~\ref{fig:supp-vis-3d}, Figure~\ref{fig:supp-vis-3d-qwen} presents direct visualizations of the feature distributions, 
where the \textit{TokenNumber} dimension is downsampled by a factor of 4 and the \textit{Dim} dimension by a factor of 2. 
Figure~\ref{fig:supp-vis-2d-1}, Figure~\ref{fig:supp-vis-2d-1-qwen} shows aggregating different tokens along the sequence dimension.
Figures~\ref{fig:supp-vis-2d-q-edit}--\ref{fig:supp-vis-2d-k-src}, Figures~\ref{fig:supp-vis-2d-q-edit-qwen}--\ref{fig:supp-vis-2d-k-src-qwen} illustrate the mean and variance of token-wise feature distributions across different attention heads, corresponding to the different embedding features.

\subsection{Kontext Embedding Visualization}
\label{subsec:kontext}

\clearpage
\subsection{Qwen-Edit Embedding Visualization}
\label{subsec:qwen}

\end{document}